\documentclass[conf]{new-aiaa}
\usepackage[utf8]{inputenc}

\usepackage{graphicx}
\usepackage{amsmath,amsfonts}
\usepackage[version=4]{mhchem}
\usepackage{siunitx}
\usepackage{longtable,tabularx}
\usepackage{booktabs}
\usepackage[table]{xcolor}
\usepackage{algorithm}
\usepackage{algorithmicx}
\usepackage{algpseudocode}
\usepackage{listings}
\usepackage{svg}
\usepackage{rotating}
\usepackage{subcaption}
\title{Surrogate-Based Optimization of System Architectures Subject to Hidden Constraints}

\author{Jasper H. Bussemaker\footnote{Researcher, DDP Group, Department of Digital Methods for System Architecting, \href{mailto:jasper.bussemaker@dlr.de}{jasper.bussemaker@dlr.de}}}
\affil{DLR (German Aerospace Center), Institute of System Architectures in Aeronautics, Hamburg, Germany}

\author{Paul Saves\footnote{Research engineer, Information Processing and Systems Department, \href{mailto:paul.saves@onera.fr}{paul.saves@onera.fr}},
Nathalie Bartoli\footnote{Senior researcher, Information Processing and Systems Department, \href{mailto:nathalie.bartoli@onera.fr}{nathalie.bartoli@onera.fr}, AIAA Member MDO TC},
Thierry Lefebvre\footnote{Research engineer, Information Processing and Systems Department, \href{mailto:thierry.lefebvre@onera.fr}{thierry.lefebvre@onera.fr}}}
\affil{DTIS, ONERA, Université de Toulouse, 31000 Toulouse, France}
\affil{Fédération ENAC ISAE-SUPAERO ONERA, Université de Toulouse, 31000, Toulouse, France}

\author{
Björn Nagel\footnote{Institute Director, Institute of System Architectures in Aeronautics, Hamburg, \href{mailto:bjoern.nagel@dlr.de}{bjoern.nagel@dlr.de}}}
\affil{DLR (German Aerospace Center), Institute of System Architectures in Aeronautics, Hamburg, Germany}

\newcommand{\noun}[1]{\textsc{#1}}

\begin{document}

\maketitle

\begin{abstract}
System Architecture Optimization (SAO) can support the design of novel architectures by formulating the architecting process as an optimization problem.
The exploration of novel architectures requires physics-based simulation due to a lack of prior experience to start from, which introduces two specific challenges for optimization algorithms: evaluations become more expensive (in time) and evaluations might fail.
The former challenge is addressed by Surrogate-Based Optimization (SBO) algorithms, in particular Bayesian Optimization (BO) using Gaussian Process (GP) models. An overview is provided of how BO can deal with challenges specific to architecture optimization, such as design variable hierarchy and multiple objectives: specific measures include ensemble infills and a hierarchical sampling algorithm.
Evaluations might fail due to non-convergence of underlying solvers or infeasible geometry in certain areas of the design space. Such failed evaluations, also known as hidden constraints, pose a particular challenge to SBO/BO, as the surrogate model cannot be trained on empty results. This work investigates various strategies for satisfying hidden constraints in BO algorithms.
Three high-level strategies are identified: \textit{rejection} of failed points from the training set, \textit{replacing} failed points based on viable (non-failed) points, and \textit{predicting} the failure region.
Through investigations on a set of test problems including a jet engine architecture optimization problem, it is shown that best performance is achieved with a mixed-discrete GP to predict the Probability of Viability (PoV), and by ensuring selected infill points satisfy some minimum PoV threshold.
This strategy is demonstrated by solving a jet engine architecture problem that features at 50\% failure rate and could not previously be solved by a BO algorithm.
% The jet engine architecture problem features a 50\% failure rate and could not previously be solved by a BO algorithm.
The developed BO algorithm and used test problems are available in the open-source Python library \noun{SBArchOpt}.
\end{abstract}
% \tableofcontents
\section*{Nomenclature}
{\renewcommand\arraystretch{1.0}
\noindent\begin{longtable*}{@{}l @{\quad=\quad} l@{}}
BO & Bayesian Optimization \\
$c_h$ & Hidden constraint \\
$c_{v,l}$ & Value constraint $l$ \\
DoE & Design of Experiments \\
FR & Failure rate \\
$f_{m}$ & Objective function $m$ \\
GP & Gaussian Process \\
$g_k$ & Inequality constraint $k$ \\
HV & Hypervolume \\
IR & Imputation ratio ($d$ subscript: discrete IR; $c$ subscript: continuous IR) \\
$k_{\mathrm{doe}}$ & DoE multiplier \\
MD & Mixed-Discrete \\
MRD & Maximum rate diversity \\
$n_{c_v}$ & Number of value constraints \\
$n_f$ & Number of objectives \\
$n_g$ & Number of inequality constraints \\
$n_{\mathrm{batch}}$ & Batch size for infill points \\
$n_{\mathrm{doe}}$ & Design of Experiments size \\
$n_{\mathrm{infill}}$ & Total number of infill points generated \\
% $n_{\mathrm{kpls}}$ & PLS components for KPLS \\
$n_{\mathrm{parallel}}$ & Maximum number of parallel evaluations \\
$n_{\mathrm{valid,discr}}$ & Number of valid discrete design vectors in an optimization problem \\
$n_x$ & Number of design variables \\
$n_{x_c}$ & Number of continuous design variables \\
$n_{x_d}$ & Number of discrete design variables \\
% $N_{fe}$ & Number of function evaluations \\
$N_j$ & Number of options for discrete variable $j$ \\
RFC & Random Forest Classifier \\
SAO & System Architecture Optimization \\
SBO & Surrogate-Based Optimization \\
$\hat{s}$ & Uncertainty estimate by a surrogate model \\
$\boldsymbol{x}$ & Design vector \\
$x_{c,i}$ & Continuous design variable $i$ \\
$x_{d,j}$ & Discrete design variable $j$ \\
$\hat{y}$ & Function value estimate by a surrogate model \\
$\delta_i$ & Activeness function for design variable $i$ \\
\end{longtable*}}

\section{Introduction}
\lettrine{S}{ystem} Architecture Optimization (SAO) is an emerging field with the goal of partial automation of designing system architectures by formulating the architecting process as a numerical optimization problem~\cite{Judt2016}. This way, a larger design space can be explored than is possible nowadays~\cite{Iacobucci2012}, and bias towards conventional solutions can be reduced~\cite{McDermott2020}.
This work deals with the aspect of developing efficient optimization algorithms for solving architecture optimization problems. For more background on architecture optimization, including how to model architecture design spaces and how to connect to evaluation code, the interested reader is referred to~\cite{Bussemaker2022c,Bussemaker2024adsg}.

\begin{figure}[h!]
\centering
\includegraphics[width=0.8\textwidth]{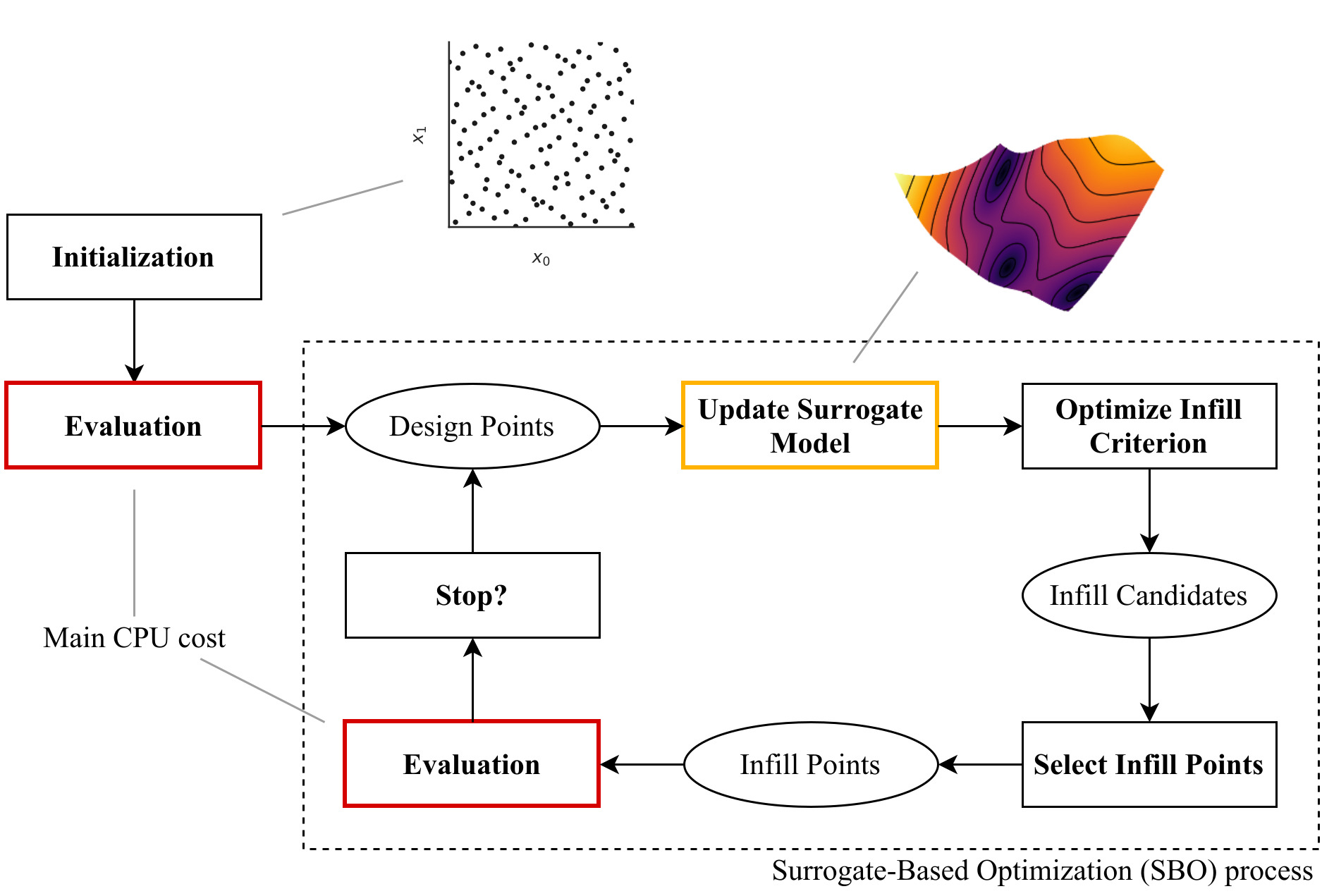}
\caption{Principle of Surrogate-Based Optimization (SBO), reproduced from~\cite{Bussemaker2024}.}\label{fig:sbo}
\end{figure}

SAO problems are subject to various aspects that make them challenging to solve: the presence of \textit{mixed-discrete} design variables stemming from architectural decisions, \textit{multiple objectives} to optimize for stemming from conflicting stakeholder needs~\cite{Crawley2015}, and the fact that the evaluation functions have to be treated as \textit{black-box} functions~\cite{Jones1998}.
Another salient feature of SAO is design variable \textit{hierarchy}~\cite{Zaefferer2018a}: the phenomenon that some design variables might decide whether other design variables are active or not, and might restrict the available options of other design variables.
Consider for example a launch vehicle design problem where both the number of rocket stages and the fuel type of each stage are design variables~\cite{Garg2024mdo}: it follows that the selection of the number of stages determines whether some of the fuel-type selection variables are active or not.
Evolutionary algorithms such as NSGA-II are well-suited for solving such optimization algorithms and are currently available for use~\cite{Judt2016,Bussemaker2021c,gamot2023hidden,Bussemaker2024}.
When searching for novel system architectures to solve for example sustainability challenges, design expertise based on historical regression analysis may be not available, however. In general when designing new architectures, physics-based simulation is needed to accurately predict the performance of architectural candidates.
This trend has two major consequences for optimization algorithms~\cite{Bussemaker2024}:
\begin{enumerate}
    \item Evaluating a design candidate is \textit{expensive}, and therefore the optimizer should aim to find the optimum (or Pareto front) is as little function evaluations as possible;
    \item Evaluation of a design candidate can \textit{fail}, for example because of unstable underlying equations, infeasible physics, or infeasible geometry. \\
\end{enumerate}
To deal with the challenges of \textit{expensive} evaluation, Surrogate Based Optimization (SBO) algorithms have been developed. These algorithms train a regression or interpolation surrogate model of the objective $f(\mathbf{x})$ and constraint $g(\mathbf{x})$ functions with respect to the design variables $\mathbf{x}$. These models are then used to generate \textit{infill} points for exploring the design space and search for the optimum or Pareto front.
The principle of SBO is visualized in Figure~\ref{fig:sbo}.
An especially powerful type of SBO is Bayesian Optimization (BO), where Gaussian Process (GP) models are used to provide not only an estimate of the function value $\hat{y}(\mathbf{x})$ but also an estimate of the associated uncertainty $\hat{s}(\mathbf{x})$.
BO is more effective at solving optimization problems than SBO algorithms with other surrogate models~\cite{GarridoMerchan2020}, and has been extended to work for constrained~\cite{Schonlau1998,Sasena2002}, mixed-discrete~\cite{GarridoMerchan2020,Munoz2020}, hierarchical~\cite{Pelamatti2020a,Audet2022,Saves2023SMT} and multi-objective~\cite{Knowles2006,RojasGonzalez2019} problems. The combination of all aspects has been demonstrated for SAO~\cite{Bussemaker2021,Bussemaker2023,Bussemaker2024}.

Evaluation failures manifest themselves as so-called \textit{hidden constraints}~\cite{Mueller2019}, which are also known as unknown, unspecified, forgotten, virtual, and crash constraints~\cite{Mueller2019,LeDigabel2023}. When a hidden constraint is violated (i.e. an evaluation has failed), the objective $f$ and constraint $g$ functions are assigned NaN (Not-a-Number).
In the taxonomy of \noun{Le Digabel \& Wild}~\cite{LeDigabel2023} hidden constraints are classified as NUSH constraints: Non-quantifiable (the degree of constraint violation is not available, only whether it is violated or not), Unrelaxable (violating the constraint yields no meaningful information about the design space), Simulation (a simulation must be run in order to find out the status), and Hidden (the existence of the constraint is not known before solving the problem).
Hidden constraints are assumed to be deterministic, mainly meaning that we do not consider evaluation failures due to temporary problems such as software license availability or network problems. Finding out whether the hidden constraint is violated may take as long, if not longer, than a non-failed evaluation~\cite{Mueller2019}.
The design space can be separated in regions with satisfied and violated hidden constraints, denoted as \textit{viable} and \textit{failed} regions, respectively.
Depending on the design problem, the failed region can take up a significant part of the design space: \noun{Krengel \& Hepperle} report up to 60\% for a wing design problem~\cite{Krengel2022}, and for an airfoil design problem \noun{Forrester et al.} up to 82\% failed points~\cite{Forrester2006}.
Hidden constraints pose a particular challenge to SBO algorithms: evaluation failures assign NaN (not-a-number) values to objective and constraints, and it is not possible to train a surrogate model on NaN's~\cite{Bussemaker2021c}.

A review of dealing with hierarchical optimization challenges for BO will be given in Section~\ref{sec:HierBO}, which is extended to hidden constraints in Section~\ref{sec:HCBO}.
The paper finishes with a demonstration of a jet engine optimization problem in Section~\ref{sec:application} and the conclusions in Section~\ref{sec:conclusions}.
All test problems used in this work are available in the open-source \noun{SBArchOpt}\footnote{\href{https://sbarchopt.readthedocs.io/}{https://sbarchopt.readthedocs.io/}} library~\cite{Bussemaker2023a}.
Code used to run the experiments is available open-source as well, at \href{https://github.com/jbussemaker/ArchitectureOptimizationExperiments/tree/hidden-constraints}{{\small github.com/jbussemaker/ArchitectureOptimizationExperiments} (\noun{hidden-constraints} branch)}.

\section{Bayesian Optimization in Hierarchical Design Spaces}\label{sec:HierBO}

Before diving into strategies for dealing with hidden constraints, this section reviews strategies for dealing with the other challenges of SAO using Bayesian Optimization (BO) algorithms as originally published in~\cite{Bussemaker2024}, and introduces the BO algorithm used in the rest of this work.
The algorithm is implemented and available as \noun{ArchSBO} in \noun{SBArchOpt}.

\paragraph{Gaussian Process Model Selection}

For non-hierarchical problems, the mixed-discrete Gaussian Process (GP) models (also known as Kriging models) implemented in the Surrogate Model Toolbox (SMT)\footnote{\href{https://smt.readthedocs.io/}{https://smt.readthedocs.io/}} are used. These models support continuous, integer and categorical variables through kernels developed by \noun{Saves et al.}~\cite{Saves2023a,Saves2023SMT}.
GP models that support hierarchical variables have been developed in~\cite{Saves2023SMT} and extended to also support categorical hierarchical variables in~\cite{Bussemaker2024}.
% Investigations have shown that these hierarchical GP models outperform non-hierarchical GP models for problems with non-linear or non-smooth objective or constraint functions.

A disadvantage of GP models is that it becomes computationally expensive to train these models for high-dimensional spaces~\cite{Priem2023}. One way of dealing with this is to define a feature space of lower dimension to train the GP, and provide a conversion between input and feature spaces~\cite{Calandra2016}.
One specific method is Kriging with Partial Least Squares (KPLS)~\cite{Bouhlel2016}, which has recently been extended to work with discrete variables~\cite{Saves2022b,Charayron2023,Saves2024} and is implemented in SMT~\cite{Saves2023SMT}.
% Investigations show that using Kriging with Partial Least Squares (KPLS)~\cite{Bouhlel2016} works well for architecture optimization problems and significantly reduces training time, at a moderate cost of optimizer performance. As a rule-of-thumb, we apply KPLS for problems with more than 10 design variables, and set the number of KPLS components $n_{\mathrm{kpls}}$ to 10 in that case.
% By default, KPLS is applied with $n_{kpls} = 10$ if the design space contains more than 10 design variables.

\paragraph{Selecting Infill Points}

Infill points are selected by optimizing infill criteria (also known as acquisition functions), and selecting the best one for a wide range of optimization problems influences BO performance significantly.
The BO algorithm uses an enable of infills, with the size of the infill batch set to the maximum number of parallel equations: $n_{\mathrm{batch}} = n_{\mathrm{parallel}}$.
% It was shown that using an ensemble of infills performs best. The size of the infill batch should be set to the maximum number of parallel equations: $n_{\mathrm{batch}} = n_{\mathrm{parallel}}$.
An infill ensemble combines multiple underlying infills into a multi-objective infill problem and thereby provides two main advantages~\cite{CowenRivers2020}:
\begin{enumerate}
    \item The selected infill points are a compromise of the underlying infills, thereby mitigating problems with different infill criteria suggesting to explore wildly different parts of the design space~\cite{Lyu2018};
    \item Since the infill problem itself results in a Pareto front, it is easy to select multiple infill points for batch evaluation without needing to retrain surrogate models~\cite{Garnett2023}. \\
\end{enumerate}

\noindent
The infill ensemble is built-up of the following underlying infills:
\begin{itemize}
    \item For single-objective optimization: Lower Confidence Bound (LCB)~\cite{Cox1992}, Expected Improvement (EI)~\cite{Jones1998}, and Probability of Improvement (PoI)~\cite{Hawe2007};
    \item For multi-objective optimization: Minimum PoI (MPoI)~\cite{Rahat2017} and Minimum Euclidean PoI (MEPoI)~\cite{Bussemaker2021}. \\
\end{itemize}

\noindent
Selecting the best infill points is an optimization problem itself, and with the same design space as the SAO problem. To successfully deal with mixed-discrete hierarchical variables, the following sequential optimization procedure is applied:
\begin{enumerate}
    \item Apply NSGA-II to search the mixed-discrete, hierarchical design space, solving the multi-objective infill problem:
    \[\begin{array}{ll}
        \mathrm{minimize} & f_{\mathrm{infill,m}}(\boldsymbol{x}_d, \boldsymbol{x}_c) \\
        \mathrm{w.r.t.} & \boldsymbol{x}_d, \boldsymbol{x}_c \\
        \mathrm{subject\ to} & \hat{g}_k(\boldsymbol{x}) \leq 0
    \end{array}\]
    where $f_{\mathrm{infill,m}}$ represents infill criterion $m$, $\boldsymbol{x}_d$ and $\boldsymbol{x}_c$ represent the discrete and continuous design variables, and $\hat{g}_k$ represents the predicted mean of constraint function $k$. Infill objectives $f_{\mathrm{infill,m}}$ are normalized and inverted such that they become minimization objectives.
    
    \item Select $n_{\mathrm{batch}}$ points $\boldsymbol{x}_{\mathrm{sel}}$ from the resulting Pareto front according to:
    \begin{itemize}
        \item If $n_{\mathrm{batch}} = 1$, select a random point;
        \item If $n_{\mathrm{batch}} > 1$, select the points with the lowest crowding distance as used in NSGA-II~\cite{Deb2011}.
    \end{itemize}
    
    \item For each selected point $\boldsymbol{x}_{sel,i}$, improve the active continuous variables by solving following single-objective problem using \noun{SLSQP}:
    \[\begin{array}{ll}
        \mathrm{minimize} & f_{\mathrm{impr}}(\boldsymbol{x}) \\
        \mathrm{w.r.t.} & \boldsymbol{x}_c \cap \delta \left( \boldsymbol{x}_{\mathrm{sel,i}} \right) \\
        \mathrm{subject\ to} & \hat{g}_k(\boldsymbol{x}) \leq 0
    \end{array}\]
    where $\boldsymbol{x}_c \cap \delta \left( \boldsymbol{x}_{\mathrm{sel,i}} \right)$ represents the active continuous design variables at point $\boldsymbol{x}_{\mathrm{sel,i}}$, and $f_{\mathrm{impr}}$ the scalar improvement objective:
    \[\begin{array}{ll}
        \Delta f_{\mathrm{infill,m}}(\boldsymbol{x}) &= f_{\mathrm{infill,m}}(\boldsymbol{x}) - f_{\mathrm{infill,m}}(\boldsymbol{x}_{sel,i}) \\
        f_{\mathrm{deviation}}(\boldsymbol{x}) &= 100 \left( \max\limits_\mathrm{m} \Delta f_{\mathrm{infill,m}}(\boldsymbol{x}) - \min\limits_\mathrm{m} \Delta f_{\mathrm{infill,m}}(\boldsymbol{x}) \right)^2 \\
        f_{\mathrm{impr}}(\boldsymbol{x}) &= \sum\limits_\mathrm{m} \left( \Delta f_{\mathrm{infill,m}}(\boldsymbol{x}) \right) + f_{\mathrm{deviation}}(\boldsymbol{x}) \\
    \end{array}\]
    This objective promotes improvement in direction of the negative unit vector, so that all underlying infill objectives are improved simultaneously while not deviating too much from the original $x_{\mathrm{sel,i}}$ to maintain batch diversity.
\end{enumerate}

% \subsection{Optimization in Hierarchical Design Spaces}
% \label{sec:bo_hier}
\paragraph{Sampling Hierarchical Design Spaces}

When sampling hierarchical design space for creating the initial Design of Experiments (DoE), care must be taken that all sections of the design space are sufficiently represented because of a possible large discrepancy between how often individual discrete design variable values appear~\cite{Crawley2015}.
For example, the discrete design vectors representing pure jet engines (vs turbofan engines) in the jet engine design problem of \noun{Bussemaker et al.}~\cite{Bussemaker2021c} only make up 0.02\% of the design space.
This discrepancy can be measured for each discrete design variable by the \textit{rate diversity} $\mathrm{RD}_j$ metric, or at the problem level by the \textit{max rate diversity} MRD, defined as:

\begin{align}
    \mathrm{Rates}_j &= \left\{ \mathrm{Rate}(j,\delta_j = 0), \mathrm{Rate}(j,0), .., \mathrm{Rate}(j,N_j-1) \right\} \\
    \mathrm{RD}_j &= \max \mathrm{Rates}_j - \min \mathrm{Rates}_j \\
    \mathrm{MRD} &= \max_{j \in 1,..,n_{x_d}} \mathrm{RD}_j
\end{align}
with $j$ the discrete design variable index, $\delta_j = 0$ denoting the case where design variable $j$ is inactive, $N_j$ the number of possible options for discrete design variable $j$, and $\mathrm{Rate}(j,\mathrm{value})$ the relative occurrence rate of that value in all valid discrete design vectors.
To ensure sufficient representation of design vectors subject to high rate diversities, a hierarchical sampling method is applied that works as follows:
\begin{enumerate}
    \item Generate all possible valid discrete design vectors $x_{\mathrm{valid,discr}}$ and obtain associated activeness information $\delta$;
    \item Group design vectors by active variables $x_{\mathrm{act}}$;
    % \item Weight each group by the number of active variables $n_{act}$;
    \item Uniformly sample discrete design vectors from each group;
    \item Assign values to active continuous variables using Sobol'~\cite{Renardy2021} sampling. \\
\end{enumerate}

\noindent
Investigations showed that this sampling method performs well for non-hierarchical problems, and performs better than random sampling for hierarchical problems.
% This was confirmed for the situations where the optimum either lies in the largest design vector group or the smallest, showing that groups of all sizes are represented sufficiently.
Refer to~\cite{Bussemaker2024} for more details.

\paragraph{Optimizing in Hierarchical Design Spaces}

\begin{figure}[h!]
\centering
\includegraphics[width=0.9\textwidth]{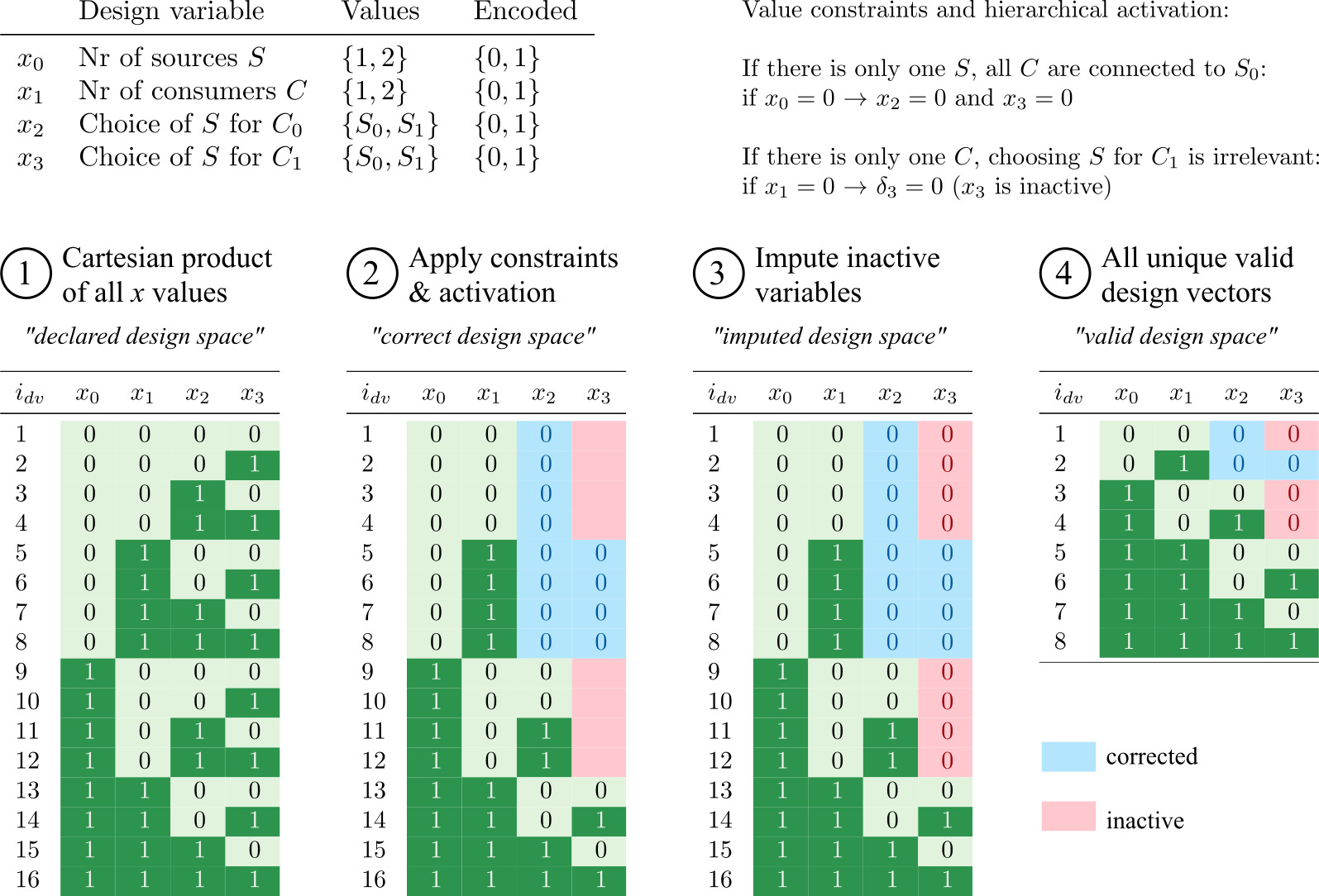}
\caption{Illustration of correction and imputation in hierarchical design spaces, showing how the different sets of design vectors relate to each other, reproduced from~\cite{Bussemaker2024}.}\label{fig:corr_imp}
\end{figure}

Finally, it was demonstrated that more integration of hierarchical information into the optimization problem yields better results. This mainly translates to the usage of the hierarchical sampling method described above, for which $x_{\mathrm{valid,discr}}$ and activeness information $\delta$ is needed, and the application of correction and imputation as a repair operator.
Correction and imputation ensure that all evaluated design vectors are valid, meaning that they satisfy value constraints and that all inactive design variables have canonical values.
Value constraints occur when selecting a value for one design variable restricts the available values for another. For example, in the Apollo mission design problem~\cite{Simmons2008}, selecting a total crew size of 2 limits the number of crew members in the lunar module to 1 or 2, whereas a total crew size of 3 means there can be 1, 2, or 3 members assigned to the lunar module.
Canonical values as applied to inactive design variables by imputation are $0$ for discrete variables and center-domain for continuous variables.
The principles of correction and imputation are visualized in Figure~\ref{fig:corr_imp}, showing how the declared design space (consisting of the Cartesian product of all discrete design variable values) is transformed to the valid design space (consisting of unique corrected and imputed design vectors).

\clearpage

Correction and imputation are made available as a standalone repair operator so that it can be used during sampling. This is important, because there might be a large difference between the declared design space as defined by the Cartesian product of all design variable values, and the valid design space containing only valid design vectors. This difference is measured by the \textit{imputation ratio} IR, defined as:

\begin{align}
    \mathrm{IR}_d &= \dfrac{\mathop{\prod_{j=1}^{n_{x_d}}N_{j}}}{n_{\mathrm{valid,discr}}} \label{eq:IRd}\\
    \mathrm{IR}_c &= \dfrac{n_{\mathrm{valid,discr}}\cdot n_{x_c}}{\mathop{\sum_{l=1}^{n_{\mathrm{valid,discr}}}}\mathop{\sum_{i=1}^{n_{x_c}}\delta_{i}(\boldsymbol{x}_{d,l})}} \label{eq:IRc}\\
    \mathrm{IR} &= \mathrm{IR}_d \cdot \mathrm{IR}_c \label{eq:IR}
\end{align}
with $n_{x_d}$ and $n_{x_c}$ the number of discrete and continuous design variables, respectively, $n_{\mathrm{valid,discr}}$ the number of valid discrete design vectors, $\delta_i$ the activeness function for design variable $i$, and $\mathrm{IR}_d$ and $\mathrm{IR}_c$ the discrete and continuous imputation ratio, respectively.
An imputation ratio of $1$ indicates no hierarchy. Higher values represent the ratio between the declared and valid design spaces sizes and thereby design space hierarchy. Imputation ratio can also be interpreted as the amount of random vectors that need to be generated in order to find one valid design vector.
For example for the suborbital vehicle design problem in~\cite{Frank2016} there are $123e3$ possible architectures, however the Cartesian product of design variables yields $2.8e6$ design vectors: therefore, the imputation ratio $\mathrm{IR} = 2.8e6 / 123e3 = 22.8$. This means that on average for every 22.8 randomly generated design vectors, there will be only one that represents a valid design.
Without correction and imputation it will be difficult for an optimization algorithm to generate new valid design vectors due to this effect. This applies to BO as well as to evolutionary optimization algorithms.
% A problem-agnostic correction algorithm is used that corrects invalid discrete design vectors to the closest valid discrete design vector, as measured by the Euclidean distance. This correction algorithm, however, depends on the availability of $x_{\mathrm{valid,discr}}$: if this is not available then problem-specific correction should be used.

\clearpage
\noindent
To conclude, the BO algorithm described in this section and implemented as \noun{ArchSBO} in \noun{SBArchOpt} features~\cite{Bussemaker2024}:
\begin{itemize}
    \item Hierarchical mixed-discrete Gaussian Process models, optionally using Kriging with Partial Least Squares (KPLS) to reduce training times for high-dimensional problems;
    \item Ensemble infill criteria with a sequential-optimization procedure for batch infill generation for single- and multi-objective optimization problems;
    % \item Sequential-optimization for selecting infill points;
    \item Hierarchical sampling algorithm that groups valid discrete design vectors by active design variables $x_{\mathrm{act}}$ to deal with rate diversity effects;
    \item Availability of correction and imputation as a repair operator to deal with imputation ratio effects.
    % \item Problem-agnostic similarity-based correction algorithm and the availability of correction and imputation as a repair operator to deal with imputation ratio effects.
\end{itemize}

\section{Hidden Constraint Strategies in Bayesian Optimization}\label{sec:HCBO}

This section identifies and investigates various strategies for dealing with hidden constraints in Bayesian Optimization (BO) algorithms.
Although we only discuss the integration with BO, the following discussions apply to any Surrogate-Based Optimization (SBO) algorithm in principle.

\subsection{Literature Review and Strategy Classification}

Hidden constraints are encountered in many engineering problems, \noun{Müller \& Day}~\cite{Mueller2019} provide several examples. They also provide an overview of strategies tried in the past, including for non-BO algorithms.
For example, for algorithms that depend on some ranking-based selection procedure rather than surrogate model building, such as evolutionary algorithms, it suffices to apply the extreme barrier approach: replace NaN by $+\infty$ to prevent these points being selected as parents for creating offspring for the next generation of design points.
% Note that this assumes that objective minimization is the goal.
Local optimization algorithms can deal with hidden constraints by implementing some way to retrace part of their search path. For SAO, however, we only consider methods relevant for global optimization algorithms.
For surrogate-based algorithms, one of the simplest methods is to train the surrogate models only on viable points, thereby in effect \textit{rejecting} failed points from the training set. The disadvantage to this approach is that knowledge of the design space is ignored: the optimizer might get stuck suggesting the same infill point(s) over and over, because it cannot know that these infill points will fail to evaluate.
A more advanced approach is to \textit{replace} the failed points by some values derived from viable points, as initially suggested by \noun{Forrester et al.}~\cite{Forrester2006} and inspired by imputation in the sense of replacing missing data in statistical datasets. They reason that failed points actually represent missing data and can be replaced by values of close-by viable points. However, the replaced values should drive the optimizer towards the viable region of the design space, which leads them to formulate a method for finding replacement values from a Gaussian Process (GP) model trained on the viable points only:
\begin{equation}
    y_{\mathrm{replace}}\left(\mathbf{x}_{\mathrm{failed}}\right) = \hat{y}\left(\mathbf{x}_{\mathrm{failed}}\right) + \alpha \cdot \hat{s}\left(\mathbf{x}_{\mathrm{failed}}\right)
\end{equation}
where $y$ represents the output value to be replaced, $\hat{y}$ and $\hat{s}$ the output and uncertainty estimates of the GP trained on viable design points, and $\alpha$ some multiplier ($\alpha = 1$ in~\cite{Forrester2006}). They show that their approach works well for a continuous single-objective airfoil optimization problem.
Another strategy for replacing values of failed points is by simply selecting one or more neighbor points and applying some aggregation function to a scalar value to replace. For example, \noun{Huyer \& Neumaier}~\cite{Huyer2008} replace failed values by the $\max$ of $n_{\mathrm{nb}}$ neighbor points. This concept can be expanded by considering $n_{\mathrm{nb}} = 1$ to use the closest point to select the replacement value, or $n_{\mathrm{nb}} = n_{\mathrm{viable}}$ to consider all viable points for the replacement value.
% Here the choice of distance function is relevant as we will be optimizing in mixed-discrete, hierarchical design spaces.

The most advanced method for dealing with hidden constraints in SBO algorithms is to \textit{predict} where the failed region lies with the help of another surrogate model~\cite{Lee2010}. Here the idea is to assign binary labels to all design point based on their failure status, for example $0$ meaning failed and $1$ meaning viable, train a model on this data, and then use that model to predict the so-called Probability of Viability (PoV).
PoV lies between 0\% and 100\% and represents the probability that a newly-selected infill point will be viable (i.e. will not fail). During infill optimization, PoV can be used in two ways: either as a penalty multiplier to the infill criterion~\cite{Lee2010}, or as a constraint to the infill algorithm that ensures that PoV of selected infill points is at or above some threshold~\cite{Alimo2018}.
Different types of surrogate models have been used to predict PoV, mainly selected due to their ability for modeling such binary classification problems. Used models include Random Forest Classifiers (RFC)~\cite{Lee2010}, piecewise linear Radial Basis Functions (RBF)~\cite{Mueller2019}, Support Vector Machines (SVM)~\cite{Sacher2018}, SVM's with RBF kernel~\cite{Alimo2018}, Gaussian Process models~\cite{Gelbart2014,Bachoc2020}, and K-Nearest Neighbors (KNN) classifiers~\cite{Audet2020,Tfaily2023}.

\clearpage
\noindent
To summarize the preceding discussion, there are three high-level strategies for dealing with hidden constraints in BO, also summarized in Table~\ref{tab:hc_strat}:
\begin{itemize}
    \item Rejection: ignore failed points when training surrogate models;
    \item Replacement: replace values of failed points based on values of viable points, either from neighboring points or by predicted replacement;
    \item Prediction: train an additional surrogate model to predict the Probability of Validity (PoV), and using this either as infill constraint or as a penalty to the infill objectives when searching for infill points.
\end{itemize}
% Replacement can be further subdivided into neighborhood replacement, taking an aggregate over various amounts of closest points, or predicted replacement, where a separate surrogate model trained on viable points is used to predict values of the failed points.
%
% Prediction can be subdivided by two aspects: the integration in the infill problem and the predictor model. Integration in the infill problem can either be done as penalty to the infill objective $f_{\mathrm{infill}}$ or as a separate infill constraint $g_{\mathrm{infill}}$ that ensures that $\mathrm{PoV}(\mathbf{x}) \geq \mathrm{PoV}_{\mathrm{min}}$. The predictor model can in principle be any binary classification model that also exposes the probability of belonging to one of the two classes in order to express PoV.

\begin{table}
\caption{Comparison of strategies for dealing with hidden constraints in Bayesian Optimization.}\label{tab:hc_strat}
\small
\centering
\begin{tabular}{llll}
\toprule
Strategy & Sub-strategy& Configuration & Parameter(s) \\
\midrule
Rejection & & & \vspace{.5cm}\\

Replacement & Neighborhood~\cite{Huyer2008} &
    $\begin{cases} \ \\ \ \\ \ \\ \ \end{cases}$ \hspace{-.5cm}
    \begin{tabular}{@{}l@{}} Closest \\ $n$-nearest, mean \\ $n$-nearest, max \\ Global, max \end{tabular}
& \begin{tabular}{@{}l@{}} $\ $ \\ $n$ \\ $n$ \\ $\ $ \end{tabular} \\
 & Predicted worst~\cite{Forrester2006} & & $\alpha$ \vspace{.5cm}\\

Prediction &
    $\begin{cases} \ \\ \ \\ \ \\ \ \\ \ \end{cases}$ \hspace{-.5cm}
    \begin{tabular}{@{}l@{}}
        Random Forest Classifier (RFC)~\cite{Lee2010} \\
        Gaussian Process (GP)~\cite{Gelbart2014} \\
        Support Vector Machine (SVM)~\cite{Sacher2018} \\
        Piecewise linear RBF~\cite{Mueller2019} \\
        K-Nearest Neighbor (KNN) classifier~\cite{Audet2020}
    \end{tabular}
    & \begin{tabular}{@{}l@{}}
        as $f_{\mathrm{infill}}$ penalty \\
        as $g_{\mathrm{infill}}$
    \end{tabular} & \begin{tabular}{@{}l@{}}
        $\ $ \\
        $\mathrm{PoV}_{\mathrm{min}}$
    \end{tabular} \\
\bottomrule
\end{tabular}
\end{table}

\subsection{Implementation into Bayesian Optimization Algorithms}

\begin{figure}[h!]
\centering
\includegraphics[width=0.63\textwidth]{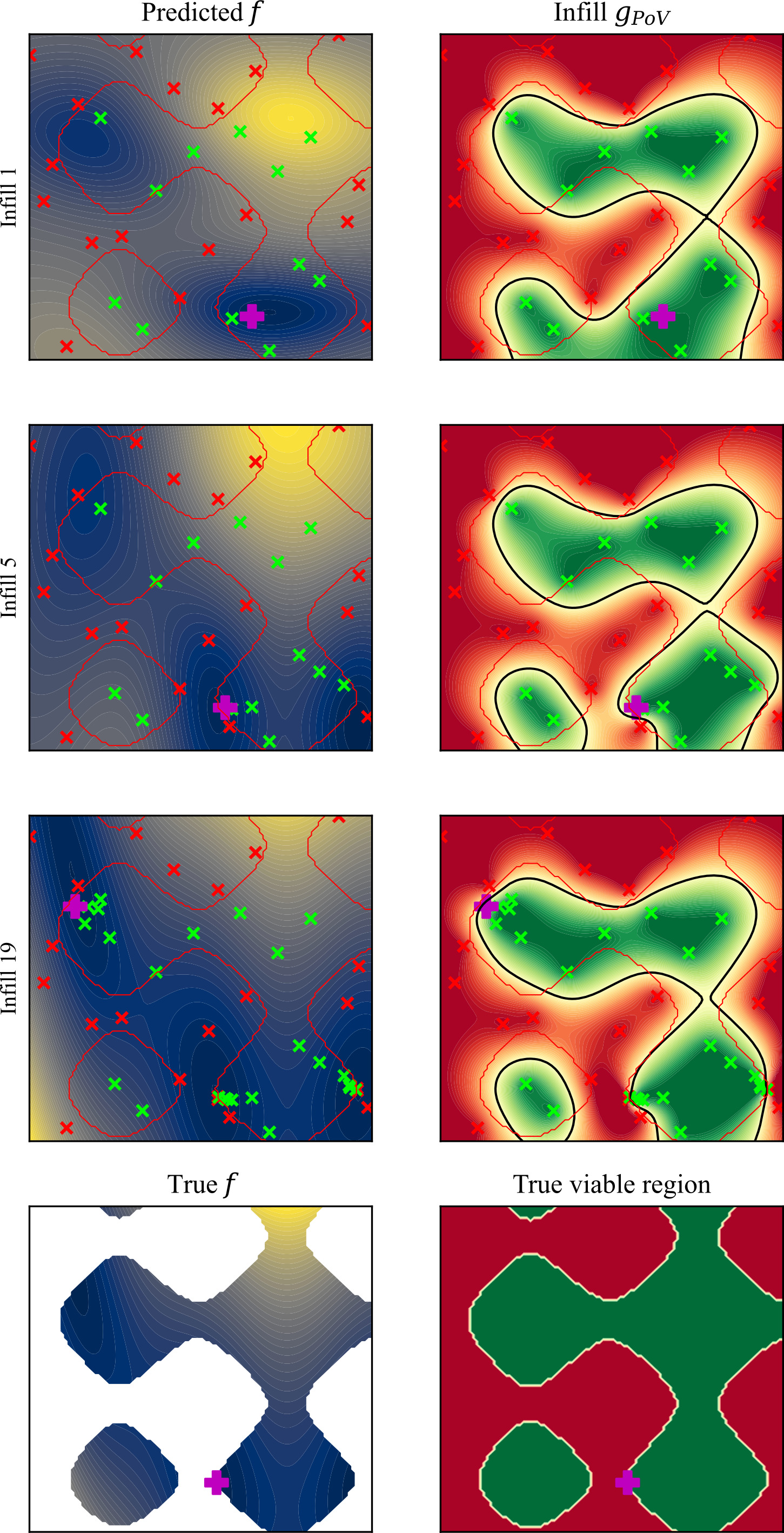}
\caption{Several optimization steps between iteration 1 and 20 of BO executed on a test problem with its optimum lying at the edge to the failed region, as shown in the bottom row by the magenta cross. The main GP is shown on the left (darker means a lower, more optimal value), and the RBF model for predicting PoV is shown on the right. The RBF model is used as an infill constraint with $\mathrm{PoV}_{\mathrm{min}} = 50\%$, showing green and red contours for satisfied and violated constraint values, respectively. Green, red, and magenta points represent viable, failed, and selected infill points, respectively.}\label{fig:hc_opt_seq}
\end{figure}

The rejection and replacement strategies are implemented into the BO algorithm in a preprocessing step before training the GP models. The training set is separated into a viable and a failed set; for rejection the viable set is then simply discarded, whereas for replacement the points in the failed set are assigned some value for each output (i.e. each objective and constraint value) that is based on the viable points. Afterwards, the GP models are trained and the infill selection process continues as usual.
For the prediction strategy, the viable set is used to train the GP models for infill search.
% The integration of the prediction strategy is more involved, as here the infill process itself is modified. As for rejection and replacement, the training set is separated into the viable and failed set. The viable set is then directly used to train the GP models for infill search.
The additional surrogate model for Probability of Viability (PoV) prediction is trained with a set containing all points and with binary labels assigned according to the viability status of the points: $0$ for failed points and $1$ for viable points.
The infill optimization problem can then be modified by adding an inequality constraint that ensures that $\mathrm{PoV}(\mathbf{x}) \geq \mathrm{PoV}_{\mathrm{min}}$:
\begin{equation}
\label{eq:g_infill}
    g_{\mathrm{PoV}}(\mathbf{x}) = \mathrm{PoV}_{\mathrm{min}} - \mathrm{PoV}(\mathbf{x}) \leq 0
\end{equation}
where $\mathrm{PoV}_{\mathrm{min}}$ represents a user-defined minimum PoV to be reached, and $\mathrm{PoV}(\mathbf{x})$ represents the predicted PoV for a given design point.
PoV can also be integrated by modifying the infill objectives:
\begin{equation}
\label{eq:f_infill}
    f_{\mathrm{m,infill,mod}}(\mathbf{x}) = 1 - \left( \left( 1 - f_{\mathrm{m,infill}}(\mathbf{x}) \right) \cdot \mathrm{PoV}(\mathbf{x}) \right)
\end{equation}
where $f_{\mathrm{m,infill}}$ represents the $m^{\mathrm{th}}$ infill objective, and assuming infill objectives are normalized and to be minimized.
Figure~\ref{fig:hc_opt_seq} presents several steps of running BO on a test problem, with an RBF model as PoV predictor used as an infill constraint at $\mathrm{PoV}_{\mathrm{min}} = 50\%$.
The test problem is the single-objective problem from \noun{Alimo et al.}~\cite{Alimo2018}, modified to have its optimum at the edge to the failed region near the top left corner of the 2D design space, implemented as \noun{AlimoEdge} in \noun{SBArchOpt}.
As can be seen, all infill points suggested satisfy the $g_{\mathrm{PoV}}$ constraint, however, in the earlier iterations this constraint can be inaccurate. Several infills are generated that violate the hidden constraint, and after each iteration the model gets more accurate at the edges to the failed regions for three locations in the design space where the optimizer expects the optimum to lie.
This visualization additionally demonstrates that the predictor model should be able to handle one or more regions in the design space with closely-spaced failed and viable points.

\clearpage

\begin{table}[t!]
\caption{Test problems for comparing hidden constraint strategies. All problems are available in the \noun{SBArchOpt} library~\cite{Bussemaker2023a}. Abbreviations and symbols: IR = imputation ratio, HC = hidden constraints, MD = mixed-discrete, H = hierarchical, MO = multi-objective, FR = fail rate, $n_{x_{c}}$ and $n_{x_{d}}$ = number of continuous and discrete design variables, respectively, $n_{f}$ = number of objectives, $n_{g}$ = number of constraints.}
\label{tab:test_problems}
\centering
\small
\begin{tabular}{lllcccccc}
\toprule
Name & \noun{SBArchOpt} Class & Ref. & $n_{x_{c}}$ & $n_{x_{d}}$ & $n_{f}$ & $n_{g}$ & IR & FR \\
\midrule
Branin & \noun{Branin} &~\cite{Forrester2008} & 2 &  & 1 &  &  & 0\% \\
HC Branin & \noun{HCBranin} &~\cite{Gelbart2014} & 2 &  & 1 & &  & 33\% \\
Alimo & \noun{Alimo} &~\cite{Alimo2018} & 2 &  & 1 &  &  & 51\% \\
Alimo Edge & \noun{AlimoEdge} & & 2 &  & 1 &  &  & 53\% \\
HC Sphere & \noun{HCSphere} &~\cite{Sacher2018} & 2 &  & 1 &  &  & 51\% \\
Müller 1 & \noun{Mueller01} &~\cite{Mueller2019} & 5 &  & 1 &  &  & 67\% \\
Müller 2 & \noun{Mueller02} &~\cite{Mueller2019} & 4 &  & 1 &  &  & 40\% \\
HC CantBeam & \noun{CantileveredBeamHC} & & 4 &  & 1 & 1 &  & 83\% \\
HC Carside Less & \noun{CarsideHCLess} & & 7 &  & 1 & 9 &  & 39\% \\
HC Carside & \noun{CarsideHC} & & 7 &  & 3 & 8 &  & 66\% \\
MD/HC CantBeam & \noun{MDCantileveredBeamHC} & & 2 & 2 & 1 & 1 &  & 81\% \\
MD/HC Carside & \noun{MDCarsideHC} & & 3 & 4 & 3 & 8 &  & 66\% \\
H Alimo & \noun{HierAlimo} & & 2 & 5 & 1 &  & 5.4 & 51\% \\
H Alimo Edge & \noun{HierAlimoEdge} & & 2 & 5 & 1 &  & 5.4 & 53\% \\
H Müller 2 & \noun{HierMueller02} & & 4 & 4 & 1 &  & 5.4 & 37\% \\
H/HC Rosenbrock & \noun{HierarchicalRosenbrockHC} &~\cite{Pelamatti2020} & 8 & 5 & 1 & 1 & 1.5 & 21\% \\
MO/H/HC Rosenbrock & \noun{MOHierarchicalRosenbrockHC} &~\cite{Pelamatti2020} & 8 & 5 & 2 & 1 & 1.5 & 60\% \\
Jet SM & \noun{SimpleTurbofanArchModel} &~\cite{Bussemaker2021c} & 9 & 6 & 1 & 5 & 3.9 & 50\% \\
\bottomrule
\end{tabular}
\end{table}

\subsection{Comparing Strategy Performances}

To compare strategy performance, we use the test problems listed in Table~\ref{tab:test_problems} to run the set of hidden constraint strategies listed in Table~\ref{tab:hc_strat_test}.
Compared to the previously identified predictor models, we additionally test a Variational GP and the mixed-discrete GP developed in~\cite{Saves2022c}.
A Variational GP does not assume a Gaussian distribution and therefore might be able to more accurately model discontinuous functions as seen in classification problems~\cite{Hensman2015}. We use the implementation provided in \noun{Trieste}\footnote{\href{https://secondmind-labs.github.io/trieste/}{https://secondmind-labs.github.io/trieste/}}~\cite{Picheny2023}.
Through a quick pre-study, we additionally discard Support Vector Machine (SVM) and piecewise linear RBF classifiers due to their bad performance in higher-dimensional and mixed-discrete test problems.
To ensure there are enough viable points to train models on for the infill search, the DoE size of problems containing hidden constraints is increased:
\begin{equation}
    n_{\mathrm{doe}} = \frac{k_{\mathrm{doe}} \cdot n_x}{1 - \mathrm{FR}_{\mathrm{exp}}}
\end{equation}
where $k_{\mathrm{doe}}$ is the DoE multiplier, $n_x$ the number of design variables, and $\mathrm{FR}_{\mathrm{exp}}$ the expected fail rate.
We assume an expected fail rate of 60\% and use $k_{\mathrm{doe}} = 2$ for following tests.
Each optimization is executed with $n_{\mathrm{infill}} = 50$ and is repeated 16 times.
Optimization performance is compared using $\Delta \mathrm{HV}$ regret, a measure that represents the distance to the known optimum or Pareto-front ($\Delta \mathrm{HV}$) cumulatively over the course of an optimization (regret). Lower $\Delta \mathrm{HV}$ regret is better, as it shows that the optimum was approached more closely and/or reached sooner.
Strategies are then ranked by the ranking procedure outlined in~\cite{Bussemaker2024}.

\clearpage
\begin{table}[h!]
\caption{Tested Bayesian Optimization hidden constraint strategies. Abbreviations: GP = Gaussian Process.}\label{tab:hc_strat_test}
% \small
\centering
\begin{tabular}{llll}
\toprule
Strategy & Sub-strategy & Configuration& Model Implementation\\
\midrule
Rejection& & &\\
Replacement& Neighborhood& Global, max&\\
& & Local&\\
& & 5-nearest, max&\\
& & 5-nearest, mean&\\
& Predicted worst& $\alpha = 1$& SMT\\%\footnotemark[1]\\
& & $\alpha = 2$& SMT\\%\footnotemark[1]\\
Prediction& Random Forest Classifier& $\mathrm{PoV}_{\mathrm{min}} = 50\%$& scikit-learn\\%\footnotemark[2] \\
& K-Nearest Neighbors& $\mathrm{PoV}_{\mathrm{min}} = 50\%$& scikit-learn\\%\footnotemark[2] \\
& Radial Basis Function& $\mathrm{PoV}_{\mathrm{min}} = 50\%$& Scipy\\%\footnotemark[3] \\
& GP Classifier& $\mathrm{PoV}_{\mathrm{min}} = 50\%$& scikit-learn\\%\footnotemark[2]\\
& Variational GP& $\mathrm{PoV}_{\mathrm{min}} = 50\%$& Trieste\\%\footnotemark[4] \\
& Mixed-Discrete GP& $\mathrm{PoV}_{\mathrm{min}} = 50\%$& SMT\\%\footnotemark[1] \\
\bottomrule
\end{tabular}
% \footnotetext[1]{\href{https://smt.readthedocs.io/}{https://smt.readthedocs.io/}}
% \footnotetext[2]{\href{https://scikit-learn.org/}{https://scikit-learn.org/}}
% \footnotetext[3]{\href{https://scipy.org/}{https://scipy.org/}}
% \footnotetext[4]{\href{https://secondmind-labs.github.io/trieste/}{https://secondmind-labs.github.io/trieste/}}
\end{table}

Tables~\ref{tab:hc_strat_rel_perf} and \ref{tab:hc_strat_res} presents optimization performance results.
It can be seen that prediction with a Random Forest Classifier (RFC) performs best (rank 1) or good (rank $\leq$ 2) more compared to other strategies, with mixed-discrete (MD) GP prediction and predicted worst replacement ($\alpha = 1$) closely following.
% Table~\ref{tab:hc_strat_rel_perf} presents some performance measures averaged over all test problems, relative to the rejection strategy.
A reduction in $\Delta \mathrm{HV}$ regret is usually combined with a reduction in failure rate: this makes sense, as a lower failure rate indicates a better capability of avoiding the failed region and therefore more resources spent on exploring the viable region.
Also shown is the change in training and infill times. Replacement strategies approximately double the training and infill cycle time, due to the fact that the trained GP models (one for each $f$ and $g$) contain the complete set of design points as training values, whereas for rejection and prediction only the viable points are used. Prediction strategies train an extra model to predict the PoV, which leads to a moderate increase in training and infill time.
% Prediction with an RFC and replacement by worst prediction both approximately add 100\% to the training and infill times to rejection; prediction with an MD GP adds 68\%.
%
% From the $\Delta \mathrm{HV}$ regret ranking and influence on training and infill times,

\begin{table}[h!]
\centering
\caption{Performance of hidden constraint strategies relative to the rejection strategy, averaged over all test problems at the end of the optimization runs. Darker color is better for $\Delta \mathrm{HV}$ regret and Fail rate; worse for time columns. Abbreviations: HV = hypervolume, RFC = Random Forest Classifier, KNN = K-Nearest Neighbors, RBF = Radial Basis Functions, GP = Gaussian Process.}\label{tab:hc_strat_rel_perf}
\begin{tabular}{llccccc}
\toprule
 Strategy & Sub-strategy & $\Delta HV$ regret & Fail rate & Training time & Infill time & Training + infill time \\
\midrule
Rejection &  & {\cellcolor[HTML]{F7FCF5}} \color[HTML]{000000} +0\% & {\cellcolor[HTML]{F7FCF5}} \color[HTML]{000000} +0\% & {\cellcolor[HTML]{FFF5F0}} \color[HTML]{000000} +0\% & {\cellcolor[HTML]{FFF5F0}} \color[HTML]{000000} +0\% & {\cellcolor[HTML]{FFF5F0}} \color[HTML]{000000} +0\% \\
Replacement & Global max & {\cellcolor[HTML]{EBF7E7}} \color[HTML]{000000} -9\% & {\cellcolor[HTML]{46AE60}} \color[HTML]{F1F1F1} -61\% & {\cellcolor[HTML]{69000D}} \color[HTML]{F1F1F1} +199\% & {\cellcolor[HTML]{FC9474}} \color[HTML]{000000} +74\% & {\cellcolor[HTML]{FC9373}} \color[HTML]{000000} +74\% \\
Replacement & Local & {\cellcolor[HTML]{E5F5E0}} \color[HTML]{000000} -13\% & {\cellcolor[HTML]{DEF2D9}} \color[HTML]{000000} -15\% & {\cellcolor[HTML]{67000D}} \color[HTML]{F1F1F1} +202\% & {\cellcolor[HTML]{FC9272}} \color[HTML]{000000} +76\% & {\cellcolor[HTML]{FC8B6B}} \color[HTML]{F1F1F1} +79\% \\
Replacement & 5-nearest, max & {\cellcolor[HTML]{B7E2B1}} \color[HTML]{000000} -30\% & {\cellcolor[HTML]{8DD08A}} \color[HTML]{000000} -43\% & {\cellcolor[HTML]{67000D}} \color[HTML]{F1F1F1} +207\% & {\cellcolor[HTML]{FC8D6D}} \color[HTML]{F1F1F1} +78\% & {\cellcolor[HTML]{FC8464}} \color[HTML]{F1F1F1} +84\% \\
Replacement & 5-nearest, mean & {\cellcolor[HTML]{A8DCA2}} \color[HTML]{000000} -35\% & {\cellcolor[HTML]{CDECC7}} \color[HTML]{000000} -23\% & {\cellcolor[HTML]{79040F}} \color[HTML]{F1F1F1} +193\% & {\cellcolor[HTML]{FC8D6D}} \color[HTML]{F1F1F1} +78\% & {\cellcolor[HTML]{FC8F6F}} \color[HTML]{000000} +77\% \\
Replacement & Predicted worst & {\cellcolor[HTML]{AEDEA7}} \color[HTML]{000000} -33\% & {\cellcolor[HTML]{7AC77B}} \color[HTML]{000000} -48\% & {\cellcolor[HTML]{7A0510}} \color[HTML]{F1F1F1} +192\% & {\cellcolor[HTML]{FC9E80}} \color[HTML]{000000} +67\% & {\cellcolor[HTML]{FC9576}} \color[HTML]{000000} +73\% \\
Replacement & Pred. worst ($\alpha = 2$) & {\cellcolor[HTML]{B7E2B1}} \color[HTML]{000000} -30\% & {\cellcolor[HTML]{68BE70}} \color[HTML]{000000} -53\% & {\cellcolor[HTML]{69000D}} \color[HTML]{F1F1F1} +199\% & {\cellcolor[HTML]{FC9373}} \color[HTML]{000000} +75\% & {\cellcolor[HTML]{FC8D6D}} \color[HTML]{F1F1F1} +78\% \\
Prediction & RFC & {\cellcolor[HTML]{8ACE88}} \color[HTML]{000000} -44\% & {\cellcolor[HTML]{48AE60}} \color[HTML]{F1F1F1} -61\% & {\cellcolor[HTML]{FC8060}} \color[HTML]{F1F1F1} +86\% & {\cellcolor[HTML]{FB7353}} \color[HTML]{F1F1F1} +94\% & {\cellcolor[HTML]{FC8565}} \color[HTML]{F1F1F1} +83\% \\
Prediction & KNN & {\cellcolor[HTML]{C2E7BB}} \color[HTML]{000000} -27\% & {\cellcolor[HTML]{9BD696}} \color[HTML]{000000} -39\% & {\cellcolor[HTML]{FCBEA5}} \color[HTML]{000000} +48\% & {\cellcolor[HTML]{FCAD90}} \color[HTML]{000000} +59\% & {\cellcolor[HTML]{FCBDA4}} \color[HTML]{000000} +49\% \\
Prediction & RBF & {\cellcolor[HTML]{A9DCA3}} \color[HTML]{000000} -35\% & {\cellcolor[HTML]{48AE60}} \color[HTML]{F1F1F1} -61\% & {\cellcolor[HTML]{ED392B}} \color[HTML]{F1F1F1} +126\% & {\cellcolor[HTML]{FC7F5F}} \color[HTML]{F1F1F1} +87\% & {\cellcolor[HTML]{FC8A6A}} \color[HTML]{F1F1F1} +80\% \\
Prediction & GP Classifier & {\cellcolor[HTML]{A2D99C}} \color[HTML]{000000} -37\% & {\cellcolor[HTML]{52B365}} \color[HTML]{F1F1F1} -58\% & {\cellcolor[HTML]{F85D42}} \color[HTML]{F1F1F1} +106\% & {\cellcolor[HTML]{C1161B}} \color[HTML]{F1F1F1} +156\% & {\cellcolor[HTML]{F44F39}} \color[HTML]{F1F1F1} +114\% \\
Prediction & Variational GP & {\cellcolor[HTML]{B0DFAA}} \color[HTML]{000000} -32\% & {\cellcolor[HTML]{4AAF61}} \color[HTML]{F1F1F1} -60\% & {\cellcolor[HTML]{FC9E80}} \color[HTML]{000000} +68\% & {\cellcolor[HTML]{67000D}} \color[HTML]{F1F1F1} +243\% & {\cellcolor[HTML]{820711}} \color[HTML]{F1F1F1} +189\% \\
Prediction & MD GP & {\cellcolor[HTML]{A0D99B}} \color[HTML]{000000} -38\% & {\cellcolor[HTML]{45AD5F}} \color[HTML]{F1F1F1} -62\% & {\cellcolor[HTML]{F34C37}} \color[HTML]{F1F1F1} +116\% & {\cellcolor[HTML]{FB7252}} \color[HTML]{F1F1F1} +95\% & {\cellcolor[HTML]{FB7858}} \color[HTML]{F1F1F1} +91\% \\
\bottomrule
\end{tabular}

\end{table}

\begin{sidewaystable}
\centering
\small
\caption{Comparison of hidden constraint strategies on various test problems, ranked by $\Delta \mathrm{HV}$ regret (lower rank / darker color is better). Best performing strategy is underlined. Abbreviations: HC = hidden constraints, MD = mixed-discrete, H = hierarchical, MO = multi-objective, RFC = Random Forest Classifier, KNN = K-Nearest Neighbors, RBF = Radial Basis Functions, GP = Gaussian Process.}\label{tab:hc_strat_res}
\begin{tabular}{llcccccccccc}
\toprule
& & Branin & HC Branin & Alimo & Alimo Edge & Müller 2 & HC Sphere & Müller 1 & HC CantB & HC Carside Less & HC Carside \\
\midrule
Rejection &  & {\cellcolor[HTML]{00441B}} \color[HTML]{F1F1F1} 1 & {\cellcolor[HTML]{98D594}} \color[HTML]{000000} 4 & {\cellcolor[HTML]{157F3B}} \color[HTML]{F1F1F1} 2 & {\cellcolor[HTML]{98D594}} \color[HTML]{000000} 4 & {\cellcolor[HTML]{00441B}} \color[HTML]{F1F1F1} 1 & {\cellcolor[HTML]{157F3B}} \color[HTML]{F1F1F1} 2 & {\cellcolor[HTML]{157F3B}} \color[HTML]{F1F1F1} 2 & {\cellcolor[HTML]{D3EECD}} \color[HTML]{000000} 5 & {\cellcolor[HTML]{F7FCF5}} \color[HTML]{000000} 6 & {\cellcolor[HTML]{F7FCF5}} \color[HTML]{000000} 6 \\
Replacement & Global max & {\cellcolor[HTML]{157F3B}} \color[HTML]{F1F1F1} 2 & {\cellcolor[HTML]{4BB062}} \color[HTML]{F1F1F1} 3 & {\cellcolor[HTML]{157F3B}} \color[HTML]{F1F1F1} 2 & {\cellcolor[HTML]{D3EECD}} \color[HTML]{000000} 5 & {\cellcolor[HTML]{00441B}} \color[HTML]{F1F1F1} 1 & {\cellcolor[HTML]{98D594}} \color[HTML]{000000} 4 & {\cellcolor[HTML]{4BB062}} \color[HTML]{F1F1F1} 3 & {\cellcolor[HTML]{157F3B}} \color[HTML]{F1F1F1} 2 & {\cellcolor[HTML]{4BB062}} \color[HTML]{F1F1F1} 3 & {\cellcolor[HTML]{98D594}} \color[HTML]{000000} 4 \\
Replacement & Local & {\cellcolor[HTML]{157F3B}} \color[HTML]{F1F1F1} 2 & {\cellcolor[HTML]{157F3B}} \color[HTML]{F1F1F1} 2 & {\cellcolor[HTML]{00441B}} \color[HTML]{F1F1F1} 1 & {\cellcolor[HTML]{98D594}} \color[HTML]{000000} 4 & {\cellcolor[HTML]{4BB062}} \color[HTML]{F1F1F1} 3 & {\cellcolor[HTML]{157F3B}} \color[HTML]{F1F1F1} 2 & {\cellcolor[HTML]{157F3B}} \color[HTML]{F1F1F1} 2 & {\cellcolor[HTML]{98D594}} \color[HTML]{000000} 4 & {\cellcolor[HTML]{D3EECD}} \color[HTML]{000000} 5 & {\cellcolor[HTML]{D3EECD}} \color[HTML]{000000} 5 \\
Replacement & 5-nearest, max & {\cellcolor[HTML]{157F3B}} \color[HTML]{F1F1F1} 2 & {\cellcolor[HTML]{157F3B}} \color[HTML]{F1F1F1} 2 & {\cellcolor[HTML]{00441B}} \color[HTML]{F1F1F1} 1 & {\cellcolor[HTML]{4BB062}} \color[HTML]{F1F1F1} 3 & {\cellcolor[HTML]{00441B}} \color[HTML]{F1F1F1} 1 & {\cellcolor[HTML]{157F3B}} \color[HTML]{F1F1F1} 2 & {\cellcolor[HTML]{00441B}} \color[HTML]{F1F1F1} 1 & {\cellcolor[HTML]{4BB062}} \color[HTML]{F1F1F1} 3 & {\cellcolor[HTML]{98D594}} \color[HTML]{000000} 4 & {\cellcolor[HTML]{4BB062}} \color[HTML]{F1F1F1} 3 \\
Replacement & 5-nearest, mean & {\cellcolor[HTML]{00441B}} \color[HTML]{F1F1F1} 1 & {\cellcolor[HTML]{157F3B}} \color[HTML]{F1F1F1} 2 & {\cellcolor[HTML]{00441B}} \color[HTML]{F1F1F1} 1 & {\cellcolor[HTML]{4BB062}} \color[HTML]{F1F1F1} 3 & {\cellcolor[HTML]{157F3B}} \color[HTML]{F1F1F1} 2 & {\cellcolor[HTML]{00441B}} \color[HTML]{F1F1F1} 1 & {\cellcolor[HTML]{00441B}} \color[HTML]{F1F1F1} 1 & {\cellcolor[HTML]{98D594}} \color[HTML]{000000} 4 & {\cellcolor[HTML]{157F3B}} \color[HTML]{F1F1F1} 2 & {\cellcolor[HTML]{157F3B}} \color[HTML]{F1F1F1} 2 \\
Replacement & Predicted worst & {\cellcolor[HTML]{157F3B}} \color[HTML]{F1F1F1} 2 & {\cellcolor[HTML]{00441B}} \color[HTML]{F1F1F1} 1 & {\cellcolor[HTML]{00441B}} \color[HTML]{F1F1F1} 1 & {\cellcolor[HTML]{98D594}} \color[HTML]{000000} 4 & {\cellcolor[HTML]{00441B}} \color[HTML]{F1F1F1} 1 & {\cellcolor[HTML]{4BB062}} \color[HTML]{F1F1F1} 3 & {\cellcolor[HTML]{00441B}} \color[HTML]{F1F1F1} 1 & {\cellcolor[HTML]{00441B}} \color[HTML]{F1F1F1} 1 & {\cellcolor[HTML]{157F3B}} \color[HTML]{F1F1F1} 2 & {\cellcolor[HTML]{157F3B}} \color[HTML]{F1F1F1} 2 \\
Replacement & Pred. worst ($\alpha = 2$) & {\cellcolor[HTML]{157F3B}} \color[HTML]{F1F1F1} 2 & {\cellcolor[HTML]{157F3B}} \color[HTML]{F1F1F1} 2 & {\cellcolor[HTML]{157F3B}} \color[HTML]{F1F1F1} 2 & {\cellcolor[HTML]{4BB062}} \color[HTML]{F1F1F1} 3 & {\cellcolor[HTML]{157F3B}} \color[HTML]{F1F1F1} 2 & {\cellcolor[HTML]{98D594}} \color[HTML]{000000} 4 & {\cellcolor[HTML]{157F3B}} \color[HTML]{F1F1F1} 2 & {\cellcolor[HTML]{00441B}} \color[HTML]{F1F1F1} 1 & {\cellcolor[HTML]{4BB062}} \color[HTML]{F1F1F1} 3 & {\cellcolor[HTML]{4BB062}} \color[HTML]{F1F1F1} 3 \\
\underline{Prediction} & \underline{RFC} & {\cellcolor[HTML]{00441B}} \color[HTML]{F1F1F1} 1 & {\cellcolor[HTML]{157F3B}} \color[HTML]{F1F1F1} 2 & {\cellcolor[HTML]{00441B}} \color[HTML]{F1F1F1} 1 & {\cellcolor[HTML]{00441B}} \color[HTML]{F1F1F1} 1 & {\cellcolor[HTML]{00441B}} \color[HTML]{F1F1F1} 1 & {\cellcolor[HTML]{157F3B}} \color[HTML]{F1F1F1} 2 & {\cellcolor[HTML]{00441B}} \color[HTML]{F1F1F1} 1 & {\cellcolor[HTML]{00441B}} \color[HTML]{F1F1F1} 1 & {\cellcolor[HTML]{4BB062}} \color[HTML]{F1F1F1} \underline{3} & {\cellcolor[HTML]{00441B}} \color[HTML]{F1F1F1} \underline{1} \\
Prediction & KNN & {\cellcolor[HTML]{157F3B}} \color[HTML]{F1F1F1} 2 & {\cellcolor[HTML]{157F3B}} \color[HTML]{F1F1F1} 2 & {\cellcolor[HTML]{00441B}} \color[HTML]{F1F1F1} 1 & {\cellcolor[HTML]{4BB062}} \color[HTML]{F1F1F1} 3 & {\cellcolor[HTML]{157F3B}} \color[HTML]{F1F1F1} 2 & {\cellcolor[HTML]{157F3B}} \color[HTML]{F1F1F1} 2 & {\cellcolor[HTML]{157F3B}} \color[HTML]{F1F1F1} 2 & {\cellcolor[HTML]{157F3B}} \color[HTML]{F1F1F1} 2 & {\cellcolor[HTML]{4BB062}} \color[HTML]{F1F1F1} 3 & {\cellcolor[HTML]{98D594}} \color[HTML]{000000} 4 \\
Prediction & RBF & {\cellcolor[HTML]{157F3B}} \color[HTML]{F1F1F1} 2 & {\cellcolor[HTML]{00441B}} \color[HTML]{F1F1F1} 1 & {\cellcolor[HTML]{00441B}} \color[HTML]{F1F1F1} 1 & {\cellcolor[HTML]{157F3B}} \color[HTML]{F1F1F1} 2 & {\cellcolor[HTML]{00441B}} \color[HTML]{F1F1F1} 1 & {\cellcolor[HTML]{4BB062}} \color[HTML]{F1F1F1} 3 & {\cellcolor[HTML]{4BB062}} \color[HTML]{F1F1F1} 3 & {\cellcolor[HTML]{00441B}} \color[HTML]{F1F1F1} 1 & {\cellcolor[HTML]{157F3B}} \color[HTML]{F1F1F1} 2 & {\cellcolor[HTML]{157F3B}} \color[HTML]{F1F1F1} 2 \\
Prediction & GP Classifier & {\cellcolor[HTML]{00441B}} \color[HTML]{F1F1F1} 1 & {\cellcolor[HTML]{00441B}} \color[HTML]{F1F1F1} 1 & {\cellcolor[HTML]{00441B}} \color[HTML]{F1F1F1} 1 & {\cellcolor[HTML]{00441B}} \color[HTML]{F1F1F1} 1 & {\cellcolor[HTML]{00441B}} \color[HTML]{F1F1F1} 1 & {\cellcolor[HTML]{157F3B}} \color[HTML]{F1F1F1} 2 & {\cellcolor[HTML]{157F3B}} \color[HTML]{F1F1F1} 2 & {\cellcolor[HTML]{00441B}} \color[HTML]{F1F1F1} 1 & {\cellcolor[HTML]{4BB062}} \color[HTML]{F1F1F1} 3 & {\cellcolor[HTML]{4BB062}} \color[HTML]{F1F1F1} 3 \\
Prediction & Variational GP & {\cellcolor[HTML]{00441B}} \color[HTML]{F1F1F1} 1 & {\cellcolor[HTML]{00441B}} \color[HTML]{F1F1F1} 1 & {\cellcolor[HTML]{00441B}} \color[HTML]{F1F1F1} 1 & {\cellcolor[HTML]{00441B}} \color[HTML]{F1F1F1} 1 & {\cellcolor[HTML]{00441B}} \color[HTML]{F1F1F1} 1 & {\cellcolor[HTML]{4BB062}} \color[HTML]{F1F1F1} 3 & {\cellcolor[HTML]{98D594}} \color[HTML]{000000} 4 & {\cellcolor[HTML]{00441B}} \color[HTML]{F1F1F1} 1 & {\cellcolor[HTML]{98D594}} \color[HTML]{000000} 4 & {\cellcolor[HTML]{157F3B}} \color[HTML]{F1F1F1} 2 \\
Prediction & MD GP & {\cellcolor[HTML]{157F3B}} \color[HTML]{F1F1F1} 2 & {\cellcolor[HTML]{00441B}} \color[HTML]{F1F1F1} 1 & {\cellcolor[HTML]{00441B}} \color[HTML]{F1F1F1} 1 & {\cellcolor[HTML]{157F3B}} \color[HTML]{F1F1F1} 2 & {\cellcolor[HTML]{00441B}} \color[HTML]{F1F1F1} 1 & {\cellcolor[HTML]{157F3B}} \color[HTML]{F1F1F1} 2 & {\cellcolor[HTML]{00441B}} \color[HTML]{F1F1F1} 1 & {\cellcolor[HTML]{157F3B}} \color[HTML]{F1F1F1} 2 & {\cellcolor[HTML]{00441B}} \color[HTML]{F1F1F1} 1 & {\cellcolor[HTML]{00441B}} \color[HTML]{F1F1F1} 1 \\
\bottomrule \\
\end{tabular}
% \hspace{15pt}
\begin{tabular}{cccccccccc}
\toprule
MD/HC CantB & MD/HC Carside & H Alimo & H Alimo Edge & H Müller 2 & H/HC Rosenbr. & MO/H/HC Rbr. & Jet SM & Rank 1 & Rank $\leq$ 2 \\
\midrule
{\cellcolor[HTML]{D3EECD}} \color[HTML]{000000} 5 & {\cellcolor[HTML]{4BB062}} \color[HTML]{F1F1F1} 3 & {\cellcolor[HTML]{98D594}} \color[HTML]{000000} 4 & {\cellcolor[HTML]{98D594}} \color[HTML]{000000} 4 & {\cellcolor[HTML]{157F3B}} \color[HTML]{F1F1F1} 2 & {\cellcolor[HTML]{4BB062}} \color[HTML]{F1F1F1} 3 & {\cellcolor[HTML]{4BB062}} \color[HTML]{F1F1F1} 3 & {\cellcolor[HTML]{98D594}} \color[HTML]{000000} 4 & {\cellcolor[HTML]{E1EDF8}} \color[HTML]{000000} 11\% & {\cellcolor[HTML]{ABD0E6}} \color[HTML]{000000} 33\% \\
{\cellcolor[HTML]{157F3B}} \color[HTML]{F1F1F1} 2 & {\cellcolor[HTML]{157F3B}} \color[HTML]{F1F1F1} 2 & {\cellcolor[HTML]{157F3B}} \color[HTML]{F1F1F1} 2 & {\cellcolor[HTML]{4BB062}} \color[HTML]{F1F1F1} 3 & {\cellcolor[HTML]{00441B}} \color[HTML]{F1F1F1} 1 & {\cellcolor[HTML]{98D594}} \color[HTML]{000000} 4 & {\cellcolor[HTML]{D3EECD}} \color[HTML]{000000} 5 & {\cellcolor[HTML]{4BB062}} \color[HTML]{F1F1F1} 3 & {\cellcolor[HTML]{E1EDF8}} \color[HTML]{000000} 11\% & {\cellcolor[HTML]{82BBDB}} \color[HTML]{000000} 44\% \\
{\cellcolor[HTML]{98D594}} \color[HTML]{000000} 4 & {\cellcolor[HTML]{4BB062}} \color[HTML]{F1F1F1} 3 & {\cellcolor[HTML]{4BB062}} \color[HTML]{F1F1F1} 3 & {\cellcolor[HTML]{4BB062}} \color[HTML]{F1F1F1} 3 & {\cellcolor[HTML]{157F3B}} \color[HTML]{F1F1F1} 2 & {\cellcolor[HTML]{157F3B}} \color[HTML]{F1F1F1} 2 & {\cellcolor[HTML]{98D594}} \color[HTML]{000000} 4 & {\cellcolor[HTML]{4BB062}} \color[HTML]{F1F1F1} 3 & {\cellcolor[HTML]{ECF4FB}} \color[HTML]{000000} 6\% & {\cellcolor[HTML]{99C7E0}} \color[HTML]{000000} 39\% \\
{\cellcolor[HTML]{4BB062}} \color[HTML]{F1F1F1} 3 & {\cellcolor[HTML]{157F3B}} \color[HTML]{F1F1F1} 2 & {\cellcolor[HTML]{157F3B}} \color[HTML]{F1F1F1} 2 & {\cellcolor[HTML]{157F3B}} \color[HTML]{F1F1F1} 2 & {\cellcolor[HTML]{00441B}} \color[HTML]{F1F1F1} 1 & {\cellcolor[HTML]{157F3B}} \color[HTML]{F1F1F1} 2 & {\cellcolor[HTML]{98D594}} \color[HTML]{000000} 4 & {\cellcolor[HTML]{157F3B}} \color[HTML]{F1F1F1} 2 & {\cellcolor[HTML]{CCDFF1}} \color[HTML]{000000} 22\% & {\cellcolor[HTML]{3787C0}} \color[HTML]{F1F1F1} 67\% \\
{\cellcolor[HTML]{4BB062}} \color[HTML]{F1F1F1} 3 & {\cellcolor[HTML]{00441B}} \color[HTML]{F1F1F1} 1 & {\cellcolor[HTML]{00441B}} \color[HTML]{F1F1F1} 1 & {\cellcolor[HTML]{157F3B}} \color[HTML]{F1F1F1} 2 & {\cellcolor[HTML]{00441B}} \color[HTML]{F1F1F1} 1 & {\cellcolor[HTML]{157F3B}} \color[HTML]{F1F1F1} 2 & {\cellcolor[HTML]{157F3B}} \color[HTML]{F1F1F1} 2 & {\cellcolor[HTML]{157F3B}} \color[HTML]{F1F1F1} 2 & {\cellcolor[HTML]{99C7E0}} \color[HTML]{000000} 39\% & {\cellcolor[HTML]{105BA4}} \color[HTML]{F1F1F1} 83\% \\
{\cellcolor[HTML]{157F3B}} \color[HTML]{F1F1F1} 2 & {\cellcolor[HTML]{00441B}} \color[HTML]{F1F1F1} 1 & {\cellcolor[HTML]{00441B}} \color[HTML]{F1F1F1} 1 & {\cellcolor[HTML]{157F3B}} \color[HTML]{F1F1F1} 2 & {\cellcolor[HTML]{00441B}} \color[HTML]{F1F1F1} 1 & {\cellcolor[HTML]{00441B}} \color[HTML]{F1F1F1} 1 & {\cellcolor[HTML]{4BB062}} \color[HTML]{F1F1F1} 3 & {\cellcolor[HTML]{00441B}} \color[HTML]{F1F1F1} 1 & {\cellcolor[HTML]{58A1CF}} \color[HTML]{F1F1F1} 56\% & {\cellcolor[HTML]{105BA4}} \color[HTML]{F1F1F1} 83\% \\
{\cellcolor[HTML]{00441B}} \color[HTML]{F1F1F1} 1 & {\cellcolor[HTML]{00441B}} \color[HTML]{F1F1F1} 1 & {\cellcolor[HTML]{00441B}} \color[HTML]{F1F1F1} 1 & {\cellcolor[HTML]{00441B}} \color[HTML]{F1F1F1} 1 & {\cellcolor[HTML]{00441B}} \color[HTML]{F1F1F1} 1 & {\cellcolor[HTML]{157F3B}} \color[HTML]{F1F1F1} 2 & {\cellcolor[HTML]{4BB062}} \color[HTML]{F1F1F1} 3 & {\cellcolor[HTML]{00441B}} \color[HTML]{F1F1F1} 1 & {\cellcolor[HTML]{99C7E0}} \color[HTML]{000000} 39\% & {\cellcolor[HTML]{2979B9}} \color[HTML]{F1F1F1} 72\% \\
{\cellcolor[HTML]{00441B}} \color[HTML]{F1F1F1} 1 & {\cellcolor[HTML]{00441B}} \color[HTML]{F1F1F1} 1 & {\cellcolor[HTML]{00441B}} \color[HTML]{F1F1F1} 1 & {\cellcolor[HTML]{157F3B}} \color[HTML]{F1F1F1} 2 & {\cellcolor[HTML]{00441B}} \color[HTML]{F1F1F1} 1 & {\cellcolor[HTML]{00441B}} \color[HTML]{F1F1F1} 1 & {\cellcolor[HTML]{00441B}} \color[HTML]{F1F1F1} 1 & {\cellcolor[HTML]{00441B}} \color[HTML]{F1F1F1} 1 & {\cellcolor[HTML]{1B69AF}} \color[HTML]{F1F1F1} \underline{78\%} & {\cellcolor[HTML]{083E81}} \color[HTML]{F1F1F1} \underline{94\%} \\
{\cellcolor[HTML]{157F3B}} \color[HTML]{F1F1F1} 2 & {\cellcolor[HTML]{157F3B}} \color[HTML]{F1F1F1} 2 & {\cellcolor[HTML]{4BB062}} \color[HTML]{F1F1F1} 3 & {\cellcolor[HTML]{98D594}} \color[HTML]{000000} 4 & {\cellcolor[HTML]{00441B}} \color[HTML]{F1F1F1} 1 & {\cellcolor[HTML]{4BB062}} \color[HTML]{F1F1F1} 3 & {\cellcolor[HTML]{00441B}} \color[HTML]{F1F1F1} 1 & {\cellcolor[HTML]{157F3B}} \color[HTML]{F1F1F1} 2 & {\cellcolor[HTML]{D6E6F4}} \color[HTML]{000000} 17\% & {\cellcolor[HTML]{3787C0}} \color[HTML]{F1F1F1} 67\% \\
{\cellcolor[HTML]{157F3B}} \color[HTML]{F1F1F1} 2 & {\cellcolor[HTML]{00441B}} \color[HTML]{F1F1F1} 1 & {\cellcolor[HTML]{157F3B}} \color[HTML]{F1F1F1} 2 & {\cellcolor[HTML]{4BB062}} \color[HTML]{F1F1F1} 3 & {\cellcolor[HTML]{00441B}} \color[HTML]{F1F1F1} 1 & {\cellcolor[HTML]{00441B}} \color[HTML]{F1F1F1} 1 & {\cellcolor[HTML]{00441B}} \color[HTML]{F1F1F1} 1 & {\cellcolor[HTML]{00441B}} \color[HTML]{F1F1F1} 1 & {\cellcolor[HTML]{6AAED6}} \color[HTML]{F1F1F1} 50\% & {\cellcolor[HTML]{105BA4}} \color[HTML]{F1F1F1} 83\% \\
{\cellcolor[HTML]{00441B}} \color[HTML]{F1F1F1} 1 & {\cellcolor[HTML]{00441B}} \color[HTML]{F1F1F1} 1 & {\cellcolor[HTML]{4BB062}} \color[HTML]{F1F1F1} 3 & {\cellcolor[HTML]{4BB062}} \color[HTML]{F1F1F1} 3 & {\cellcolor[HTML]{00441B}} \color[HTML]{F1F1F1} 1 & {\cellcolor[HTML]{157F3B}} \color[HTML]{F1F1F1} 2 & {\cellcolor[HTML]{00441B}} \color[HTML]{F1F1F1} 1 & {\cellcolor[HTML]{00441B}} \color[HTML]{F1F1F1} 1 & {\cellcolor[HTML]{4695C8}} \color[HTML]{F1F1F1} 61\% & {\cellcolor[HTML]{1B69AF}} \color[HTML]{F1F1F1} 78\% \\
{\cellcolor[HTML]{00441B}} \color[HTML]{F1F1F1} 1 & {\cellcolor[HTML]{00441B}} \color[HTML]{F1F1F1} 1 & {\cellcolor[HTML]{157F3B}} \color[HTML]{F1F1F1} 2 & {\cellcolor[HTML]{4BB062}} \color[HTML]{F1F1F1} 3 & {\cellcolor[HTML]{00441B}} \color[HTML]{F1F1F1} 1 & {\cellcolor[HTML]{157F3B}} \color[HTML]{F1F1F1} 2 & {\cellcolor[HTML]{157F3B}} \color[HTML]{F1F1F1} 2 & {\cellcolor[HTML]{00441B}} \color[HTML]{F1F1F1} 1 & {\cellcolor[HTML]{58A1CF}} \color[HTML]{F1F1F1} 56\% & {\cellcolor[HTML]{1B69AF}} \color[HTML]{F1F1F1} 78\% \\
{\cellcolor[HTML]{00441B}} \color[HTML]{F1F1F1} 1 & {\cellcolor[HTML]{00441B}} \color[HTML]{F1F1F1} 1 & {\cellcolor[HTML]{00441B}} \color[HTML]{F1F1F1} 1 & {\cellcolor[HTML]{00441B}} \color[HTML]{F1F1F1} 1 & {\cellcolor[HTML]{00441B}} \color[HTML]{F1F1F1} 1 & {\cellcolor[HTML]{00441B}} \color[HTML]{F1F1F1} 1 & {\cellcolor[HTML]{D3EECD}} \color[HTML]{000000} 5 & {\cellcolor[HTML]{00441B}} \color[HTML]{F1F1F1} 1 & {\cellcolor[HTML]{2979B9}} \color[HTML]{F1F1F1} 72\% & {\cellcolor[HTML]{083E81}} \color[HTML]{F1F1F1} 94\% \\
\bottomrule
\end{tabular}

\end{sidewaystable}

\clearpage

Predicted worst replacement, prediction with RFC, and prediction with MD GP are selected as most promising candidates.
These three strategies are further investigated to find out the influence of their parameter settings: $\alpha$ for predicted worst replacement, and $\mathrm{PoV}_{\mathrm{min}}$ for the prediction strategies. In addition, the prediction strategies are tested with integration as $f$-infill penalty (Eq.~(\ref{eq:f_infill})).
Tests are run with the same settings as the previous experiment.

\begin{figure}[hb]
    \centering
    \begin{subfigure}[b]{0.55\textwidth}
        \centering
        \includegraphics[width=0.9\textwidth]{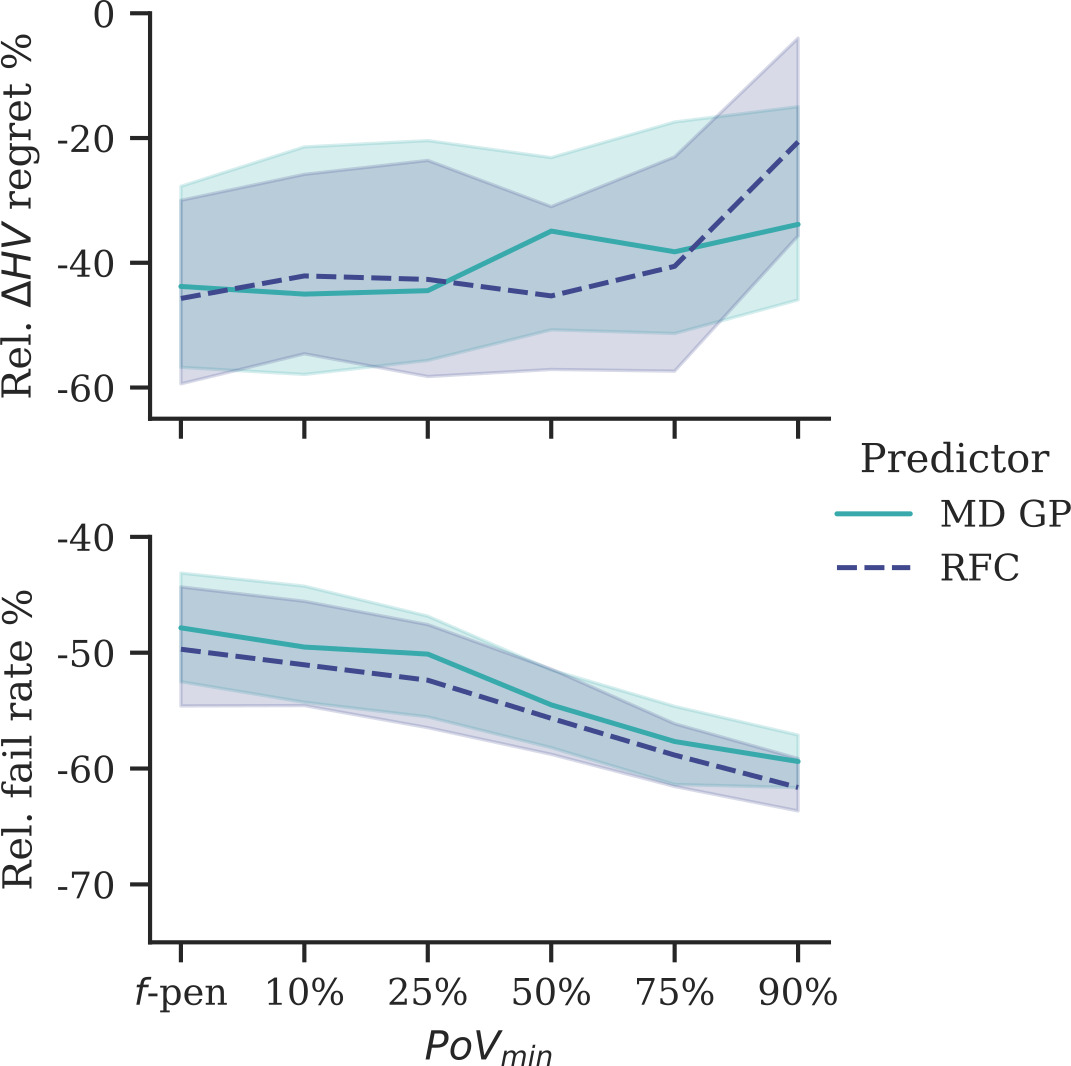}
        \caption{Predictor strategies}
    \end{subfigure}
    \hfill
    \begin{subfigure}[b]{0.42\textwidth}
        \centering
        \includegraphics[width=0.9\textwidth]{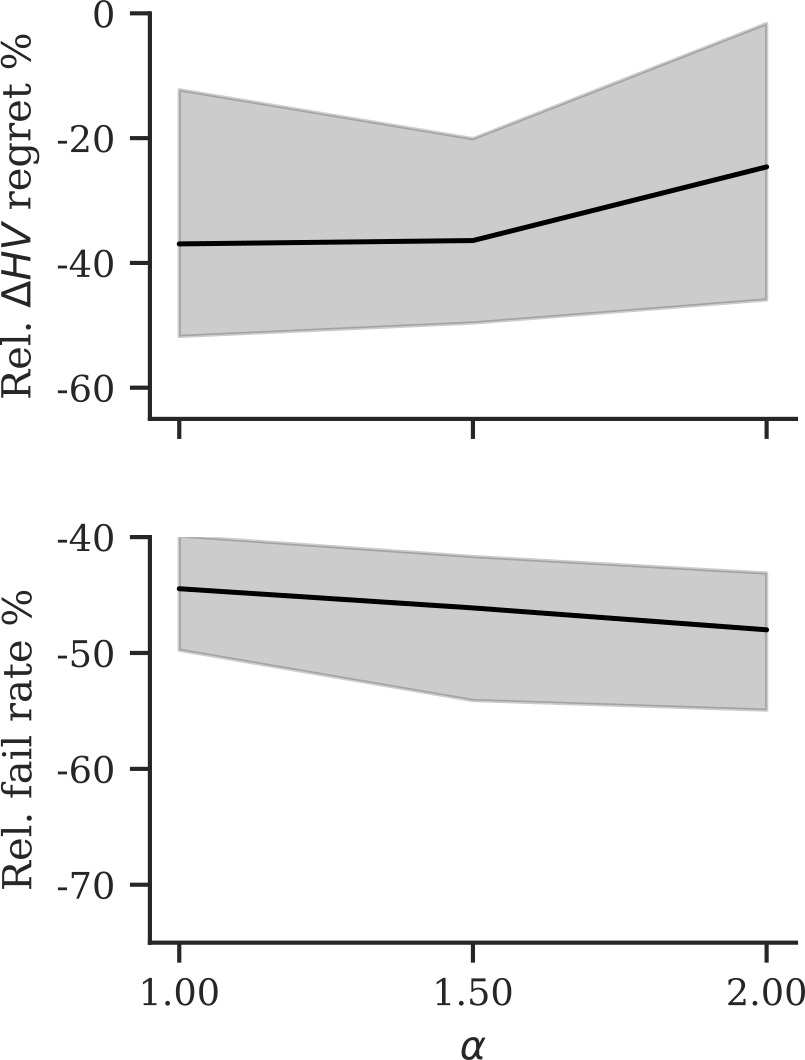}
        \caption{Predicted worst replacement strategy}
    \end{subfigure}
    \caption{Comparison of hidden constraint strategy settings relative to the rejection strategy, averaged over all test problems at the end of the optimization runs. Abbreviations: HV = hypervolume, MD = mixed-discrete, GP = Gaussian Process, RFC = Random Forest Classifier.}\label{fig:hc_strat_config_rel_perf}
\end{figure}

Figure~\ref{fig:hc_strat_config_rel_perf} shows the relative improvement over the rejection strategy and Table~\ref{tab:hc_strat_config_res} presents ranking of algorithm performance for the tested strategy configurations.
Fail rate is reduced significantly for higher values of $\mathrm{PoV}_{\mathrm{min}}$ and $\alpha$, which is expected as higher values result in a more conservative approach and therefore less exploration of the failed region. A reduction in failure rate, however, decreases optimizer performance (seen by an increase in $\Delta \mathrm{HV}$ regret), showing that a certain amount of failed evaluations is needed to sufficiently explore the design space.
The best performing strategies are the prediction strategies at low $\mathrm{PoV}_{\mathrm{min}}$ values.
It also shows that $f$-infill penalty behaves similar as low $\mathrm{PoV}_{\mathrm{min}}$ values.
Prediction with MD GP or RFC are behaving similarly-well, although MD GP for a little wider range of $\mathrm{PoV}_{\mathrm{min}}$ than RFC.
From these results, we recommend that either MD GP or RFC prediction should be used to deal with hidden constraints in Bayesian Optimization.
PoV should be integrated as a constraint, because it allows more control over exploration vs exploitation compared to integration as $f$-infill penalty.
$\mathrm{PoV}_{\mathrm{min}}$ should be kept relatively low to promote sufficient exploration: a value of $\mathrm{PoV}_{\mathrm{min}} = 25\%$ will be used subsequently.

\vfill
\vfill

\clearpage
\begin{sidewaystable}
\centering
\small
\caption{Comparison of hidden constraint strategy settings on various test problems, ranked by $\Delta \mathrm{HV}$ regret (lower rank / darker color is better). Best performing strategy is underlined. Abbreviations: HC = hidden constraints, MD = mixed-discrete, H = hierarchical, MO = multi-objective, RFC = Random Forest Classifier, KNN = K-Nearest Neighbors, RBF = Radial Basis Functions, GP = Gaussian Process.}\label{tab:hc_strat_config_res}
\begin{tabular}{llccccccccccc}
\toprule
 && Alimo & Alimo Edge & Müller 1 & MD/HC Carside & H Alimo & H/HC Rbr. & MO/H/HC Rbr. & Jet SM & Rank 1 & Rank $\leq$ 2 & Penalty \\
\midrule
Predicted Worst & $\alpha = 1.00$ & {\cellcolor[HTML]{00441B}} \color[HTML]{F1F1F1} 1 & {\cellcolor[HTML]{AEDEA7}} \color[HTML]{000000} 3 & {\cellcolor[HTML]{37A055}} \color[HTML]{F1F1F1} 2 & {\cellcolor[HTML]{AEDEA7}} \color[HTML]{000000} 3 & {\cellcolor[HTML]{00441B}} \color[HTML]{F1F1F1} 1 & {\cellcolor[HTML]{37A055}} \color[HTML]{F1F1F1} 2 & {\cellcolor[HTML]{37A055}} \color[HTML]{F1F1F1} 2 & {\cellcolor[HTML]{37A055}} \color[HTML]{F1F1F1} 2 & {\cellcolor[HTML]{C6DBEF}} \color[HTML]{000000} 25\% & {\cellcolor[HTML]{2070B4}} \color[HTML]{F1F1F1} 75\% & {\cellcolor[HTML]{2A924A}} \color[HTML]{F1F1F1} 15\% \\
Predicted Worst & $\alpha = 1.50$ & {\cellcolor[HTML]{00441B}} \color[HTML]{F1F1F1} 1 & {\cellcolor[HTML]{AEDEA7}} \color[HTML]{000000} 3 & {\cellcolor[HTML]{37A055}} \color[HTML]{F1F1F1} 2 & {\cellcolor[HTML]{37A055}} \color[HTML]{F1F1F1} 2 & {\cellcolor[HTML]{00441B}} \color[HTML]{F1F1F1} 1 & {\cellcolor[HTML]{00441B}} \color[HTML]{F1F1F1} 1 & {\cellcolor[HTML]{37A055}} \color[HTML]{F1F1F1} 2 & {\cellcolor[HTML]{37A055}} \color[HTML]{F1F1F1} 2 & {\cellcolor[HTML]{9DCAE1}} \color[HTML]{000000} 38\% & {\cellcolor[HTML]{08509B}} \color[HTML]{F1F1F1} 88\% & {\cellcolor[HTML]{218944}} \color[HTML]{F1F1F1} 13\% \\
Predicted Worst & $\alpha = 2.00$ & {\cellcolor[HTML]{37A055}} \color[HTML]{F1F1F1} 2 & {\cellcolor[HTML]{F7FCF5}} \color[HTML]{000000} 4 & {\cellcolor[HTML]{00441B}} \color[HTML]{F1F1F1} 1 & {\cellcolor[HTML]{AEDEA7}} \color[HTML]{000000} 3 & {\cellcolor[HTML]{37A055}} \color[HTML]{F1F1F1} 2 & {\cellcolor[HTML]{37A055}} \color[HTML]{F1F1F1} 2 & {\cellcolor[HTML]{AEDEA7}} \color[HTML]{000000} 3 & {\cellcolor[HTML]{37A055}} \color[HTML]{F1F1F1} 2 & {\cellcolor[HTML]{DEEBF7}} \color[HTML]{000000} 12\% & {\cellcolor[HTML]{4191C6}} \color[HTML]{F1F1F1} 62\% & {\cellcolor[HTML]{B0DFAA}} \color[HTML]{000000} 36\% \\
\underline{MD-GP} & \underline{$f_{infill}$ penalty} & {\cellcolor[HTML]{00441B}} \color[HTML]{F1F1F1} 1 & {\cellcolor[HTML]{00441B}} \color[HTML]{F1F1F1} 1 & {\cellcolor[HTML]{00441B}} \color[HTML]{F1F1F1} 1 & {\cellcolor[HTML]{00441B}} \color[HTML]{F1F1F1} 1 & {\cellcolor[HTML]{37A055}} \color[HTML]{F1F1F1} 2 & {\cellcolor[HTML]{37A055}} \color[HTML]{F1F1F1} 2 & {\cellcolor[HTML]{00441B}} \color[HTML]{F1F1F1} 1 & {\cellcolor[HTML]{00441B}} \color[HTML]{F1F1F1} 1 & {\cellcolor[HTML]{2070B4}} \color[HTML]{F1F1F1} \underline{75\%} & {\cellcolor[HTML]{08306B}} \color[HTML]{F1F1F1} \underline{100\%} & {\cellcolor[HTML]{00491D}} \color[HTML]{F1F1F1} \underline{1\%} \\
MD-GP & $PoV_{min} = 10\%$ & {\cellcolor[HTML]{00441B}} \color[HTML]{F1F1F1} 1 & {\cellcolor[HTML]{37A055}} \color[HTML]{F1F1F1} 2 & {\cellcolor[HTML]{00441B}} \color[HTML]{F1F1F1} 1 & {\cellcolor[HTML]{00441B}} \color[HTML]{F1F1F1} 1 & {\cellcolor[HTML]{37A055}} \color[HTML]{F1F1F1} 2 & {\cellcolor[HTML]{37A055}} \color[HTML]{F1F1F1} 2 & {\cellcolor[HTML]{00441B}} \color[HTML]{F1F1F1} 1 & {\cellcolor[HTML]{00441B}} \color[HTML]{F1F1F1} 1 & {\cellcolor[HTML]{4191C6}} \color[HTML]{F1F1F1} 62\% & {\cellcolor[HTML]{08306B}} \color[HTML]{F1F1F1} 100\% & {\cellcolor[HTML]{005723}} \color[HTML]{F1F1F1} 3\% \\
\underline{MD-GP} & \underline{$PoV_{min} = 25\%$} & {\cellcolor[HTML]{00441B}} \color[HTML]{F1F1F1} 1 & {\cellcolor[HTML]{37A055}} \color[HTML]{F1F1F1} 2 & {\cellcolor[HTML]{00441B}} \color[HTML]{F1F1F1} 1 & {\cellcolor[HTML]{00441B}} \color[HTML]{F1F1F1} 1 & {\cellcolor[HTML]{00441B}} \color[HTML]{F1F1F1} 1 & {\cellcolor[HTML]{37A055}} \color[HTML]{F1F1F1} 2 & {\cellcolor[HTML]{00441B}} \color[HTML]{F1F1F1} 1 & {\cellcolor[HTML]{00441B}} \color[HTML]{F1F1F1} 1 & {\cellcolor[HTML]{2070B4}} \color[HTML]{F1F1F1} \underline{75\%} & {\cellcolor[HTML]{08306B}} \color[HTML]{F1F1F1} \underline{100\%} & {\cellcolor[HTML]{00441B}} \color[HTML]{F1F1F1} \underline{-1\%} \\
MD-GP & $PoV_{min} = 50\%$ & {\cellcolor[HTML]{00441B}} \color[HTML]{F1F1F1} 1 & {\cellcolor[HTML]{37A055}} \color[HTML]{F1F1F1} 2 & {\cellcolor[HTML]{00441B}} \color[HTML]{F1F1F1} 1 & {\cellcolor[HTML]{00441B}} \color[HTML]{F1F1F1} 1 & {\cellcolor[HTML]{00441B}} \color[HTML]{F1F1F1} 1 & {\cellcolor[HTML]{37A055}} \color[HTML]{F1F1F1} 2 & {\cellcolor[HTML]{F7FCF5}} \color[HTML]{000000} 4 & {\cellcolor[HTML]{00441B}} \color[HTML]{F1F1F1} 1 & {\cellcolor[HTML]{4191C6}} \color[HTML]{F1F1F1} 62\% & {\cellcolor[HTML]{08509B}} \color[HTML]{F1F1F1} 88\% & {\cellcolor[HTML]{0C7735}} \color[HTML]{F1F1F1} 9\% \\
MD-GP & $PoV_{min} = 75\%$ & {\cellcolor[HTML]{00441B}} \color[HTML]{F1F1F1} 1 & {\cellcolor[HTML]{37A055}} \color[HTML]{F1F1F1} 2 & {\cellcolor[HTML]{37A055}} \color[HTML]{F1F1F1} 2 & {\cellcolor[HTML]{00441B}} \color[HTML]{F1F1F1} 1 & {\cellcolor[HTML]{00441B}} \color[HTML]{F1F1F1} 1 & {\cellcolor[HTML]{37A055}} \color[HTML]{F1F1F1} 2 & {\cellcolor[HTML]{AEDEA7}} \color[HTML]{000000} 3 & {\cellcolor[HTML]{00441B}} \color[HTML]{F1F1F1} 1 & {\cellcolor[HTML]{6AAED6}} \color[HTML]{F1F1F1} 50\% & {\cellcolor[HTML]{08509B}} \color[HTML]{F1F1F1} 88\% & {\cellcolor[HTML]{026F2E}} \color[HTML]{F1F1F1} 7\% \\
MD-GP & $PoV_{min} = 90\%$ & {\cellcolor[HTML]{00441B}} \color[HTML]{F1F1F1} 1 & {\cellcolor[HTML]{37A055}} \color[HTML]{F1F1F1} 2 & {\cellcolor[HTML]{37A055}} \color[HTML]{F1F1F1} 2 & {\cellcolor[HTML]{00441B}} \color[HTML]{F1F1F1} 1 & {\cellcolor[HTML]{00441B}} \color[HTML]{F1F1F1} 1 & {\cellcolor[HTML]{37A055}} \color[HTML]{F1F1F1} 2 & {\cellcolor[HTML]{F7FCF5}} \color[HTML]{000000} 4 & {\cellcolor[HTML]{00441B}} \color[HTML]{F1F1F1} 1 & {\cellcolor[HTML]{6AAED6}} \color[HTML]{F1F1F1} 50\% & {\cellcolor[HTML]{08509B}} \color[HTML]{F1F1F1} 88\% & {\cellcolor[HTML]{208843}} \color[HTML]{F1F1F1} 13\% \\
RFC & $f_{infill}$ penalty & {\cellcolor[HTML]{00441B}} \color[HTML]{F1F1F1} 1 & {\cellcolor[HTML]{00441B}} \color[HTML]{F1F1F1} 1 & {\cellcolor[HTML]{37A055}} \color[HTML]{F1F1F1} 2 & {\cellcolor[HTML]{AEDEA7}} \color[HTML]{000000} 3 & {\cellcolor[HTML]{37A055}} \color[HTML]{F1F1F1} 2 & {\cellcolor[HTML]{00441B}} \color[HTML]{F1F1F1} 1 & {\cellcolor[HTML]{00441B}} \color[HTML]{F1F1F1} 1 & {\cellcolor[HTML]{00441B}} \color[HTML]{F1F1F1} 1 & {\cellcolor[HTML]{4191C6}} \color[HTML]{F1F1F1} 62\% & {\cellcolor[HTML]{08509B}} \color[HTML]{F1F1F1} 88\% & {\cellcolor[HTML]{00441B}} \color[HTML]{F1F1F1} -2\% \\
RFC & $PoV_{min} = 10\%$ & {\cellcolor[HTML]{00441B}} \color[HTML]{F1F1F1} 1 & {\cellcolor[HTML]{00441B}} \color[HTML]{F1F1F1} 1 & {\cellcolor[HTML]{AEDEA7}} \color[HTML]{000000} 3 & {\cellcolor[HTML]{37A055}} \color[HTML]{F1F1F1} 2 & {\cellcolor[HTML]{37A055}} \color[HTML]{F1F1F1} 2 & {\cellcolor[HTML]{37A055}} \color[HTML]{F1F1F1} 2 & {\cellcolor[HTML]{00441B}} \color[HTML]{F1F1F1} 1 & {\cellcolor[HTML]{37A055}} \color[HTML]{F1F1F1} 2 & {\cellcolor[HTML]{9DCAE1}} \color[HTML]{000000} 38\% & {\cellcolor[HTML]{08509B}} \color[HTML]{F1F1F1} 88\% & {\cellcolor[HTML]{107A37}} \color[HTML]{F1F1F1} 10\% \\
RFC & $PoV_{min} = 25\%$ & {\cellcolor[HTML]{00441B}} \color[HTML]{F1F1F1} 1 & {\cellcolor[HTML]{00441B}} \color[HTML]{F1F1F1} 1 & {\cellcolor[HTML]{37A055}} \color[HTML]{F1F1F1} 2 & {\cellcolor[HTML]{37A055}} \color[HTML]{F1F1F1} 2 & {\cellcolor[HTML]{37A055}} \color[HTML]{F1F1F1} 2 & {\cellcolor[HTML]{37A055}} \color[HTML]{F1F1F1} 2 & {\cellcolor[HTML]{00441B}} \color[HTML]{F1F1F1} 1 & {\cellcolor[HTML]{00441B}} \color[HTML]{F1F1F1} 1 & {\cellcolor[HTML]{6AAED6}} \color[HTML]{F1F1F1} 50\% & {\cellcolor[HTML]{08306B}} \color[HTML]{F1F1F1} 100\% & {\cellcolor[HTML]{006729}} \color[HTML]{F1F1F1} 6\% \\
RFC & $PoV_{min} = 50\%$ & {\cellcolor[HTML]{00441B}} \color[HTML]{F1F1F1} 1 & {\cellcolor[HTML]{37A055}} \color[HTML]{F1F1F1} 2 & {\cellcolor[HTML]{37A055}} \color[HTML]{F1F1F1} 2 & {\cellcolor[HTML]{AEDEA7}} \color[HTML]{000000} 3 & {\cellcolor[HTML]{37A055}} \color[HTML]{F1F1F1} 2 & {\cellcolor[HTML]{00441B}} \color[HTML]{F1F1F1} 1 & {\cellcolor[HTML]{00441B}} \color[HTML]{F1F1F1} 1 & {\cellcolor[HTML]{00441B}} \color[HTML]{F1F1F1} 1 & {\cellcolor[HTML]{6AAED6}} \color[HTML]{F1F1F1} 50\% & {\cellcolor[HTML]{08509B}} \color[HTML]{F1F1F1} 88\% & {\cellcolor[HTML]{004C1E}} \color[HTML]{F1F1F1} 1\% \\
RFC & $PoV_{min} = 75\%$ & {\cellcolor[HTML]{00441B}} \color[HTML]{F1F1F1} 1 & {\cellcolor[HTML]{37A055}} \color[HTML]{F1F1F1} 2 & {\cellcolor[HTML]{37A055}} \color[HTML]{F1F1F1} 2 & {\cellcolor[HTML]{00441B}} \color[HTML]{F1F1F1} 1 & {\cellcolor[HTML]{37A055}} \color[HTML]{F1F1F1} 2 & {\cellcolor[HTML]{00441B}} \color[HTML]{F1F1F1} 1 & {\cellcolor[HTML]{37A055}} \color[HTML]{F1F1F1} 2 & {\cellcolor[HTML]{37A055}} \color[HTML]{F1F1F1} 2 & {\cellcolor[HTML]{9DCAE1}} \color[HTML]{000000} 38\% & {\cellcolor[HTML]{08306B}} \color[HTML]{F1F1F1} 100\% & {\cellcolor[HTML]{0E7936}} \color[HTML]{F1F1F1} 9\% \\
RFC & $PoV_{min} = 90\%$ & {\cellcolor[HTML]{00441B}} \color[HTML]{F1F1F1} 1 & {\cellcolor[HTML]{AEDEA7}} \color[HTML]{000000} 3 & {\cellcolor[HTML]{F7FCF5}} \color[HTML]{000000} 4 & {\cellcolor[HTML]{F7FCF5}} \color[HTML]{000000} 4 & {\cellcolor[HTML]{37A055}} \color[HTML]{F1F1F1} 2 & {\cellcolor[HTML]{37A055}} \color[HTML]{F1F1F1} 2 & {\cellcolor[HTML]{F7FCF5}} \color[HTML]{000000} 4 & {\cellcolor[HTML]{AEDEA7}} \color[HTML]{000000} 3 & {\cellcolor[HTML]{DEEBF7}} \color[HTML]{000000} 12\% & {\cellcolor[HTML]{9DCAE1}} \color[HTML]{000000} 38\% & {\cellcolor[HTML]{F7FCF5}} \color[HTML]{000000} 54\% \\
\bottomrule
\end{tabular}

\end{sidewaystable}

\clearpage
\section{Application: Jet Engine Architecture Optimization}\label{sec:application}

To demonstrate the hidden constraint optimization strategies presented in this work, the open-source jet engine optimization problem presented in~\cite{Bussemaker2021c} is used.
The problem contains several architectural choices, such as whether to include a fan and bypass flow, the number of stages, the addition of a gearbox, and where to apply bleed and power offtakes. Each generated engine architecture is evaluated using a pyCycle~\cite{Hendricks2019} OpenMDAO~\cite{Gray2019} problem. The Multidisciplinary Design Optimization (MDO) problem is automatically constructed from a class-based engine architecture definition. %, easing the integration with architecture generation platforms.
Next to thermodynamic cycle analysis disciplines, additional disciplines include weight estimation, length and diameter calculation, and noise and NOx estimations.

We solve the simple version of the problem, which is a single-objective minimization of Thrust-Specific Fuel Consumption (TSFC) with following architectural choices: fan inclusion, number of compressor stages, gearbox inclusion, mixed nozzle selection, and power offtake locations.
There are 70 valid discrete design vectors. However, the Cartesian product of discrete variables leads to 216 combinations: the discrete imputation ratio therefore is $\mathrm{IR}_d = 216 / 70 = 3.1$ (see Eq.~\eqref{eq:IRd}).
The continuous imputation ratio $\mathrm{IR}_c = 1.26$ (see Eq.~\eqref{eq:IRc}), meaning that there are on average $9 / 1.26 = 7.14$ continuous variables active (as seen over all valid discrete design vectors).
The overall imputation ratio is $\mathrm{IR} = 3.89$ (see Eq.~\eqref{eq:IR}).
Additionally, the underlying thermodynamic cycle analysis and sizing code does not always converge, leading to a hidden constraint being violated in approximately 50\% of design points generated in a random DoE.
This version of the problem is implemented in \noun{SBArchOpt} as \noun{SimpleTurbofanArch}.

\begin{figure}[h!]
\centering
\includegraphics[width=0.75\textwidth]{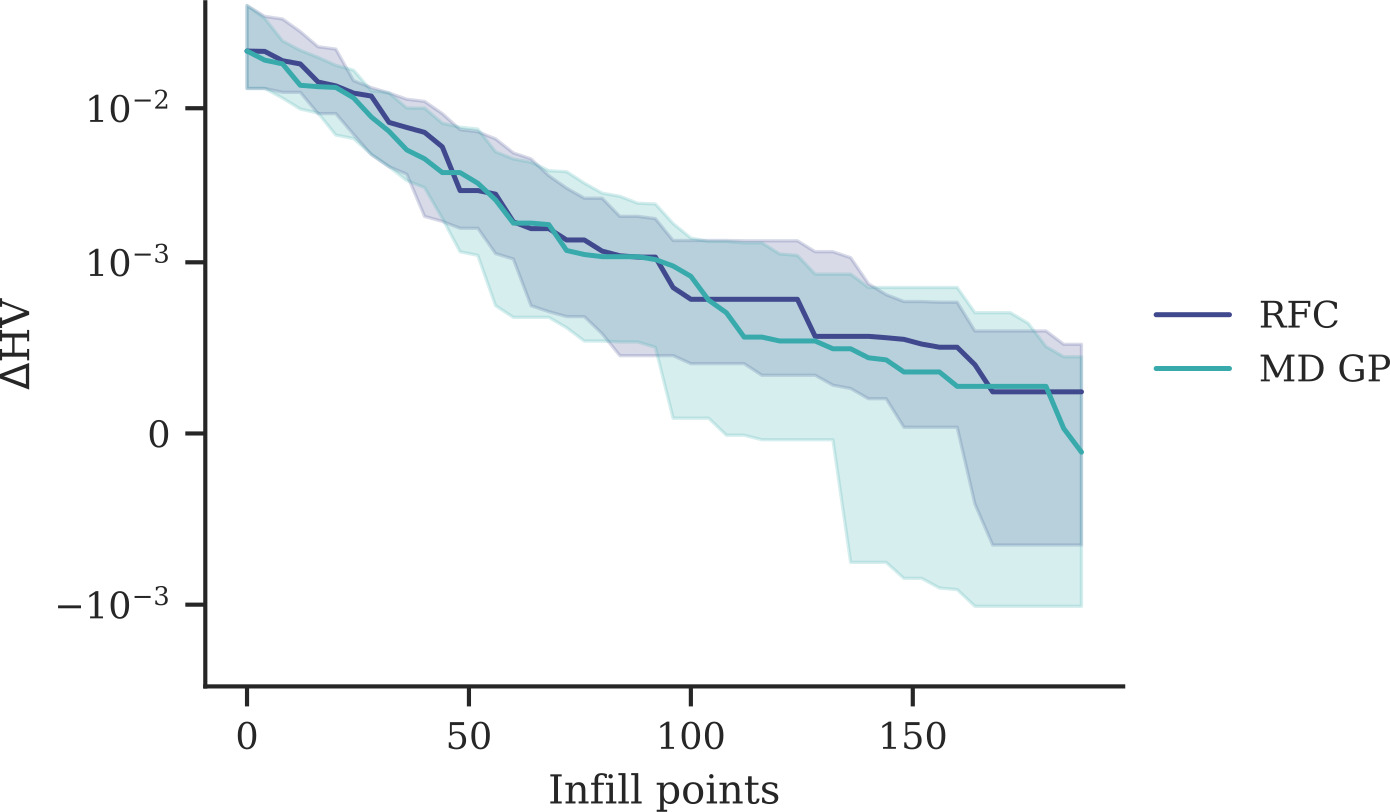}
\caption{Optimization progression of the BO algorithm with the prediction hidden constraint strategy with two different predictors for the jet engine optimization problem.
$\Delta \mathrm{HV}$ represents the distance to the known Pareto front.
The bands around the lines represent the 50 percentile range around the median, obtained from 24 repeated runs.
Abbreviations: MD = mixed-discrete, GP = Gaussian Process, RFC = Random Forest Classifier.}\label{fig:jet_eng_opt}
\end{figure}

The BO algorithm is executed 24 times with $n_{\mathrm{doe}} = 113$, $n_{\mathrm{infill}} = 187$ (a budget of 300 evaluations), and $n_{\mathrm{batch}} = 4$. Both hidden constraint strategies use $\mathrm{PoV}_{\mathrm{min}} = 25\%$.
Figure~\ref{fig:jet_eng_opt} presents the results of the jet engine optimization.
It shows that the RFC and MD GP predictors perform similarly, with the MD GP predictor slightly outperforming, and both being able to find the known optimum within the 300 evaluations.
The known optimum was found with NSGA-II and an evaluation budget of 3250~\cite{Bussemaker2024}; BO therefore can be considered to be able to find the same result in 92\% less function evaluations.
% The MD GP predictor clearly outperforms the RFC predictor: it progresses quicker towards the known optimum and finds a better optimum value after the 140 infill points.
% The optimum has been found in 75\% less function evaluations compared to NSGA-II, where a budget of 1000 evaluations was used in~\cite{Bussemaker2021c}.
% The MD GP predictor clearly outperforms the RFC predictor: it progresses quicker towards the known optimum and finds a better optimum value after the 400 infill points.
% The optimum has been found in 77\% less function evaluations compared to NSGA-II, where a budget of 4000 evaluations was used in~\cite{Bussemaker2021c}.

% \vfill
% \vfill

\clearpage
\section{Conclusions and Outlook}\label{sec:conclusions}

System Architecture Optimization (SAO) problems are challenging to solve due to their mixed-discrete and hierarchical design spaces combined with constrained, multi-objective, black-box, and expensive-to-evaluate solution spaces that potentially are subject to hidden constraints.
Bayesian Optimization (BO) algorithms are gradient-free, global optimization algorithms that are specifically developed to optimize with a small evaluation budget.
A BO algorithm is presented that has been developed for optimization in hierarchical design spaces, featuring:
\begin{itemize}
    \item Hierarchical mixed-discrete Gaussian Process models, optionally using Kriging with Partial Least Squares (KPLS) to reduce training times for high-dimensional problems;
    \item Ensemble infill criteria with a sequential-optimization procedure for batch infill generation for single- and multi-objective optimization problems;
    \item Hierarchical sampling algorithm that groups valid discrete design vectors by active design variables $x_{\mathrm{act}}$ to deal with rate diversity effects;
    \item Availability of correction and imputation as a repair operator to deal with imputation ratio effects.\\
\end{itemize}

\noindent
The presented algorithm is extended to deal with hidden constraints.
First, three high-level strategies are identified: rejection (ignore failed points), replacement (replace output values of failed points by some value derived from viable points), and prediction (predict the Probability of Viability).
Replacement strategies are subdivided into replacement from neighboring points and replacement by prediction.
Prediction strategies are subdivided based on the used classification models.
It is demonstrated how prediction strategies can be implemented in BO, either as a penalty to the infill objectives or as additional inequality constraints to the infill problem parameterized by a lower threshold $\mathrm{PoV}_{\mathrm{min}}$.

Then, multiple hidden constraint strategies are tested on a database of test problems.
It is shown that predicted worst replacement, prediction with a Random Forest Classifier (RFC), and prediction with a mixed-discrete Gaussian Process (GP) perform best.
Compared to rejection, training and infill times are increased due to the increase in training points for replacement strategies and training of the PoV prediction model for prediction strategies.
These three strategies are further tested with different values of $\mathrm{PoV}_{\mathrm{min}}$ (for prediction) and $\alpha$ for predicted worst replacement. It is shown that prediction with MD GP or RFC performs best, and with lower $\mathrm{PoV}_{\mathrm{min}}$ values. A default value of $\mathrm{PoV}_{\mathrm{min}} = 25\%$ is recommended.
Finally, a jet engine architecture optimization problem is solved with the two predictor strategies, showing that prediction with MD GP performs best although it only slightly outperforms the RFC predictor.
The extended BO algorithm and all used test problems are available in the open-source \noun{SBArchOpt} library.

This work has shown that hidden constraints can be effectively dealt with by BO algorithms in the context of SAO.
To validate its applicability, BO should be applied to more aerospace and non-aerospace architecture optimization problems.
In future work the use of BO for SAO should be expanded, specifically to larger design space sizes with tens to hundreds of design variables, for example using dimension reduction techniques such as EGORSE~\cite{Priem2023}, and multi-fidelity BO should be considered as a way to make BO more effective and further reduce required computational resources.
% ~\cite{Donelli2023}

% To support users in applying the developed algorithms for SAO, it should be easy to select, configure, and execute optimizations directly from system architecting tools.
% For example, architecture problems formulated using ADORE~\cite{Bussemaker2024adsg} could be translated to an \noun{SBArchOpt} problem definition without additional user effort. From there, any algorithm implemented in \noun{SBArchOpt} could be selected and executed to solve the defined optimization problem.

% \appendix
% \section*{Appendix}
% \subsection*{Overview of Used Test Problems}\label{sec:test_problems}

\section*{Acknowledgments}
The research presented in this paper has been performed in the framework of the COLOSSUS project (Collaborative System of Systems Exploration of Aviation Products, Services and Business Models) and has received funding from the European Union Horizon Europe Programme under grant agreement n${^\circ}$ 101097120.
The authors would like to recognize Luca Boggero and Björn Nagel for supporting the secondment of Jasper Bussemaker at ONERA DTIS in Toulouse, during which the research presented in this paper was mostly produced.

\bibliography{main} %,library}

\end{document}